\documentclass{article} 
\usepackage{iclr2027_conference,times}
\usepackage{bbm}
\usepackage{makecell}
\usepackage{multirow}
\usepackage{amssymb}
\usepackage{amsmath}
\usepackage{booktabs}
\usepackage{subfiles}
\usepackage{float}
\usepackage{subcaption}
\usepackage{algorithm}
\usepackage{algpseudocode}
\usepackage{tabularx}
\usepackage{graphicx}
\usepackage{wrapfig}
\usepackage{hyperref}
\usepackage{url}
\usepackage[table]{xcolor}
\definecolor{gainrow}{RGB}{220,235,252}
\definecolor{gainred}{RGB}{220,0,0}

\usepackage{amsmath,amsfonts,bm}

\def\eqref#1{equation~\ref{#1}}

\def\1{\bm{1}}

\DeclareMathAlphabet{\mathsfit}{\encodingdefault}{\sfdefault}{m}{sl}
\SetMathAlphabet{\mathsfit}{bold}{\encodingdefault}{\sfdefault}{bx}{n}

\usepackage{hyperref}
\usepackage{url}
\iclrfinalcopy

\title{PF-RL: Progress Field Reinforcement Learning via Goal-Conditioned Value Geometry for Vision-Language-Action Models}

\author{%
  Yunpeng Qing$^{1}$, Yilun Kong$^{2}$,
  Sixu Lin$^{3}$, Ming Zhou$^{4}$,
  Yiming Fei$^{1}$, Shuang Luo$^{2}$,\\
  \textbf{Yixiao Chi$^{5}$, Haoming Gu$^{6}$,
  Jingyuan Liu$^{7}$, Changxu Wei$^{8}$,
  Zhi Hou$^{9}$, Changqing Zou$^{1,10}$}  \vspace{3pt}
    \\
  $^{1}$ Zhejiang University, $^{2}$  Nanyang Technological University, $^{3}$ CUHK-Shenzhen,\\
  $^{4}$ Shanghai AI Laboratory,
  $^{5}$ Carnegie Mellon University,
  $^{6}$ Harbin Institute of Technology,\\
  $^{7}$ Fudan University, $^{8}$ PolyU,
  $^{9}$ ACE Robotics, $^{10}$ Zhejiang Lab
}
\begin{document}

\maketitle
\fancyhead{}
\lhead{Preprint}
\begin{abstract}
  Reinforcement Fine-Tuning~(RFT) has emerged as a promising paradigm for improving Vision-Language-Action~(VLA) policies, yet sparse task-level outcomes provide limited credit for intermediate transitions, especially in long-horizon manipulation.
  A natural approach is to model intermediate task progress and use it as dense feedback for policy improvement.
  Despite their architectural differences, existing progress-aware methods commonly formulate task progress as an explicit scalar prediction, providing limited structure for modeling how intermediate observations relate to the task goal, which may hinder effective transition-level credit assignment.
  We introduce \textbf{Progress Field Reinforcement Learning~(PF-RL)}, which learns a structured goal-conditioned progress representation over pretrained VLA features and converts it into dense credit for policy optimization.
  A lightweight shared Progress Field head maps current and goal representations into a compact progress space, where geometric distance induces goal-conditioned value, while complementary temporal and goal-structure objectives shape the learned geometry.
  Transition-level value changes naturally yield dense progress advantages, enabling fine-grained credit assignment for both offline policy improvement and online reinforcement fine-tuning. 
  Extensive experiments on LIBERO, RoboTwin2.0, and real-world bimanual manipulation tasks show that PF-RL consistently improves policy performance over strong supervised fine-tuning, reinforcement fine-tuning, and progress-aware baselines.
\end{abstract}
\section{Introduction}

Vision-Language-Action~(VLA) models have recently shown remarkable potential for general-purpose robotic manipulation, benefiting from large-scale multimodal pretraining~\citep{team2026ace} and imitation learning~\citep{kim2024openvla,intelligence2025pi_}. 
Beyond supervised fine-tuning, Reinforcement Learning~(RL)~\citep{qing2024a2po,qing2025bitrajdiff,liu2025curricular} has emerged as a promising post-training paradigm for VLA models, enabling further policy improvement through interaction and task feedback~\citep{lu2025vla,chen2025tgrpo,li2026simplevla,kong2026extoken,lin2026dygro}.
Despite these advances, current VLA-RL pipelines still rely heavily on sparse task-level outcomes, such as binary success or failure.
This becomes particularly limiting in long-horizon manipulation, where a single terminal signal must provide learning feedback for many intermediate transitions that contribute differently to task completion.
Such coarse trajectory-level supervision reveals whether an episode ultimately succeeds, but provides little guidance about which intermediate transitions advance the task and which lead to stagnation or regression.
Consequently, credit assignment remains a key bottleneck for Reinforcement Fine-Tuning~(RFT) of VLA policies.

\begin{figure*}[t]
    \centering

    \begin{subfigure}[t]{0.26\textwidth}
        \centering
        \includegraphics[width=\linewidth]
        {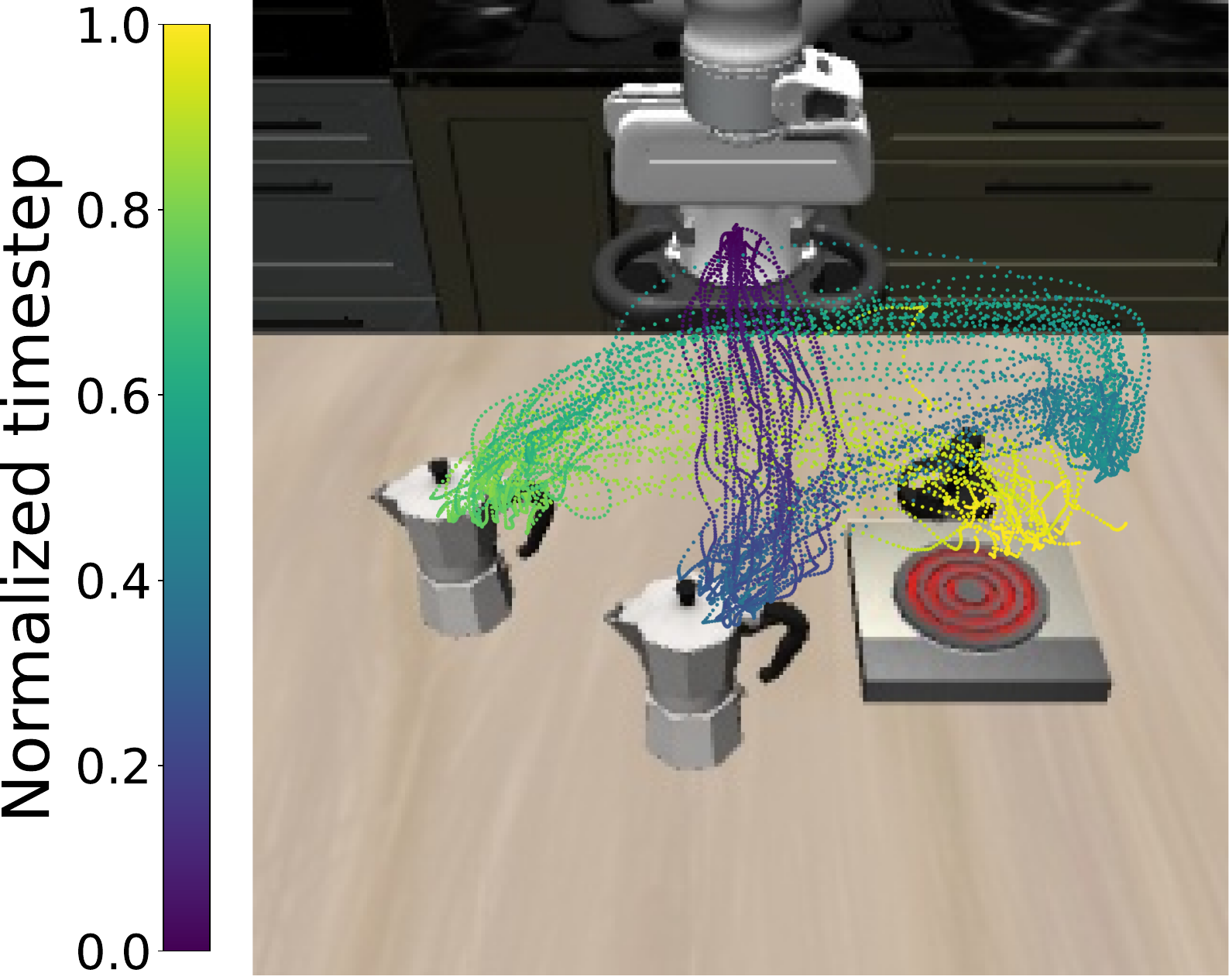}
        \caption{Task progression}
        \label{fig:motivation_traj}
    \end{subfigure}
    \hfill
    \begin{subfigure}[t]{0.23\textwidth}
        \centering
        \includegraphics[width=\linewidth]
        {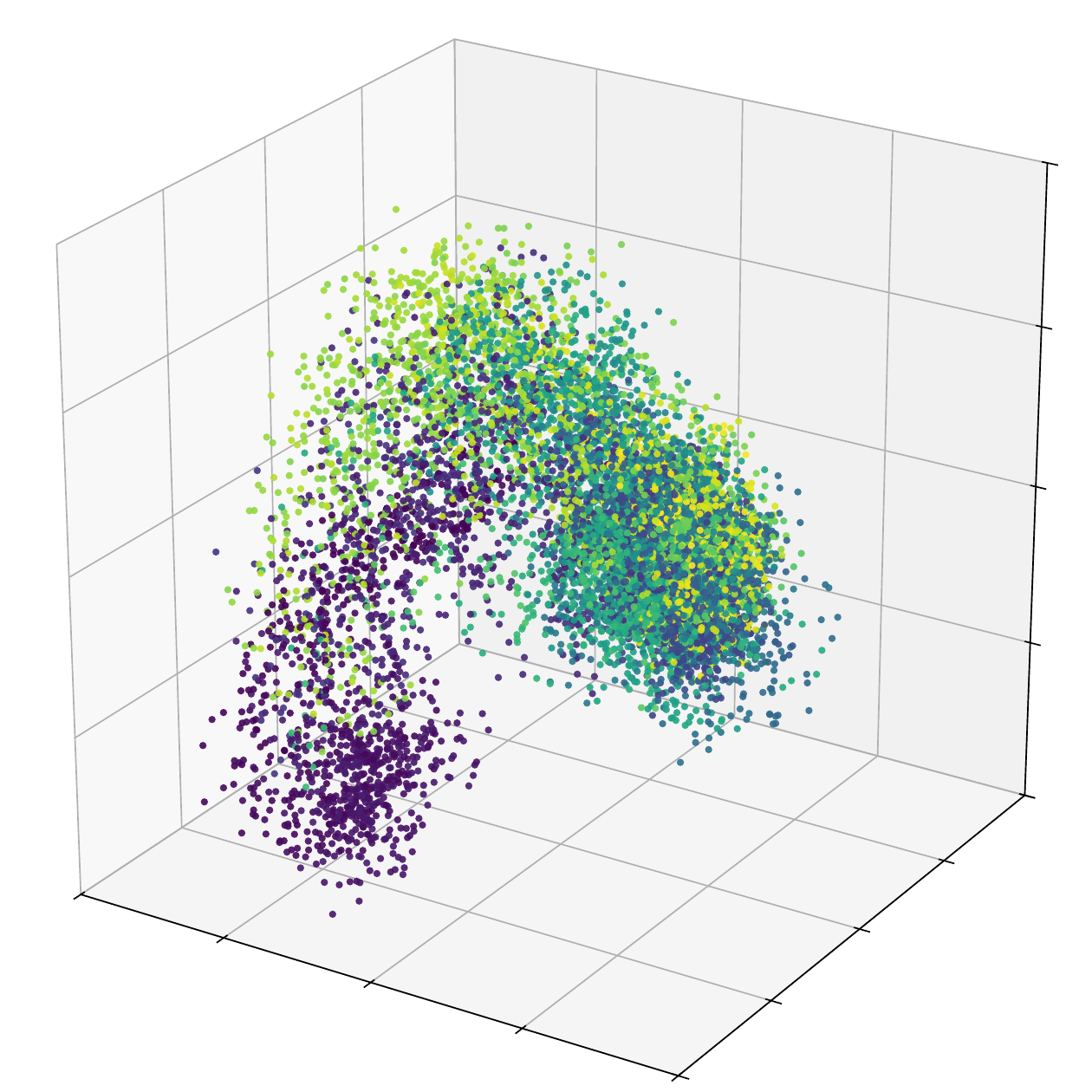}
        \caption{Raw VLA space}
        \label{fig:motivation_raw}
    \end{subfigure}
    \hfill
    \begin{subfigure}[t]{0.23\textwidth}
        \centering
        \includegraphics[width=\linewidth]
        {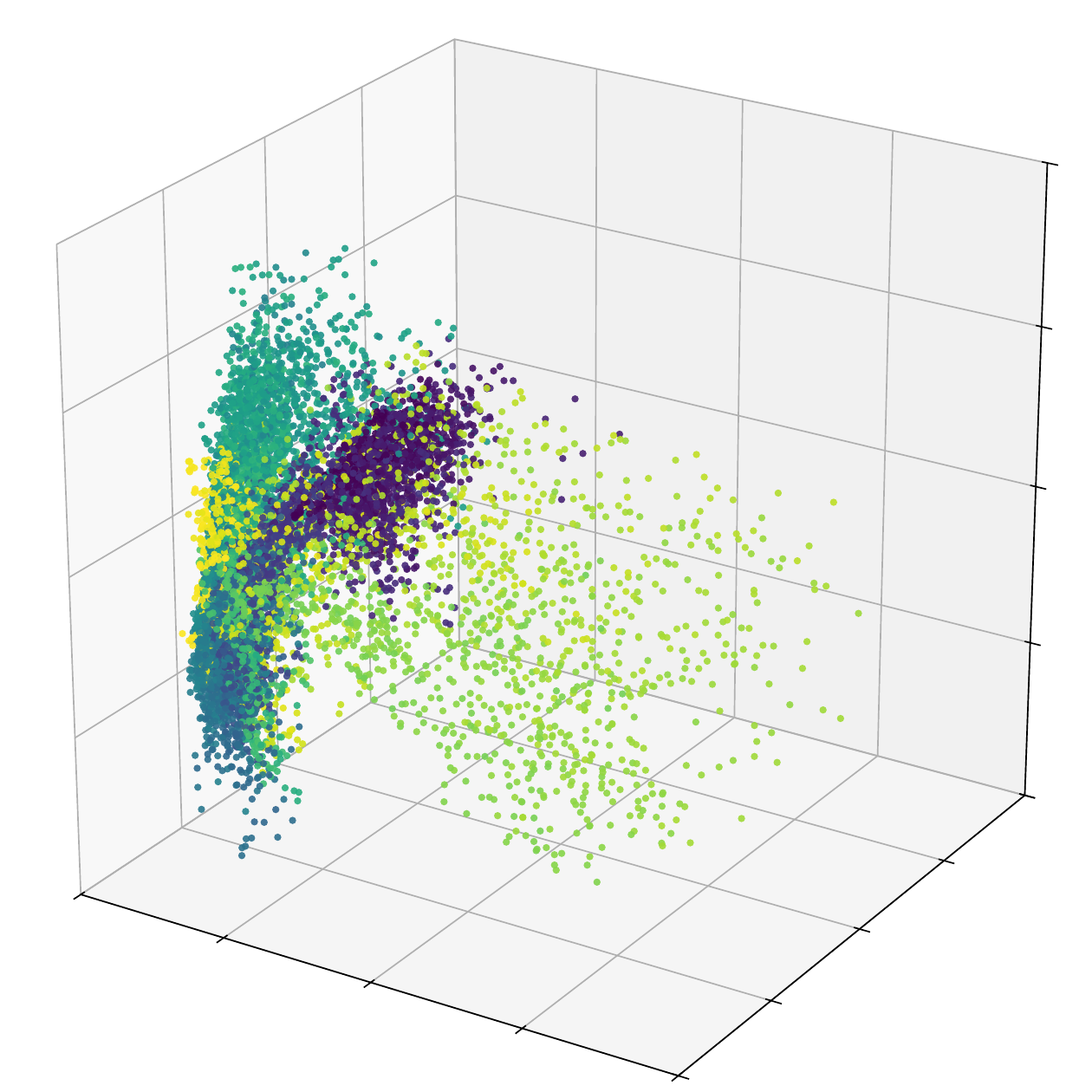}
        \caption{Scalar prediction}
        \label{fig:motivation_scalar}
    \end{subfigure}
    \hfill
    \begin{subfigure}[t]{0.23\textwidth}
        \centering
        \includegraphics[width=\linewidth]
        {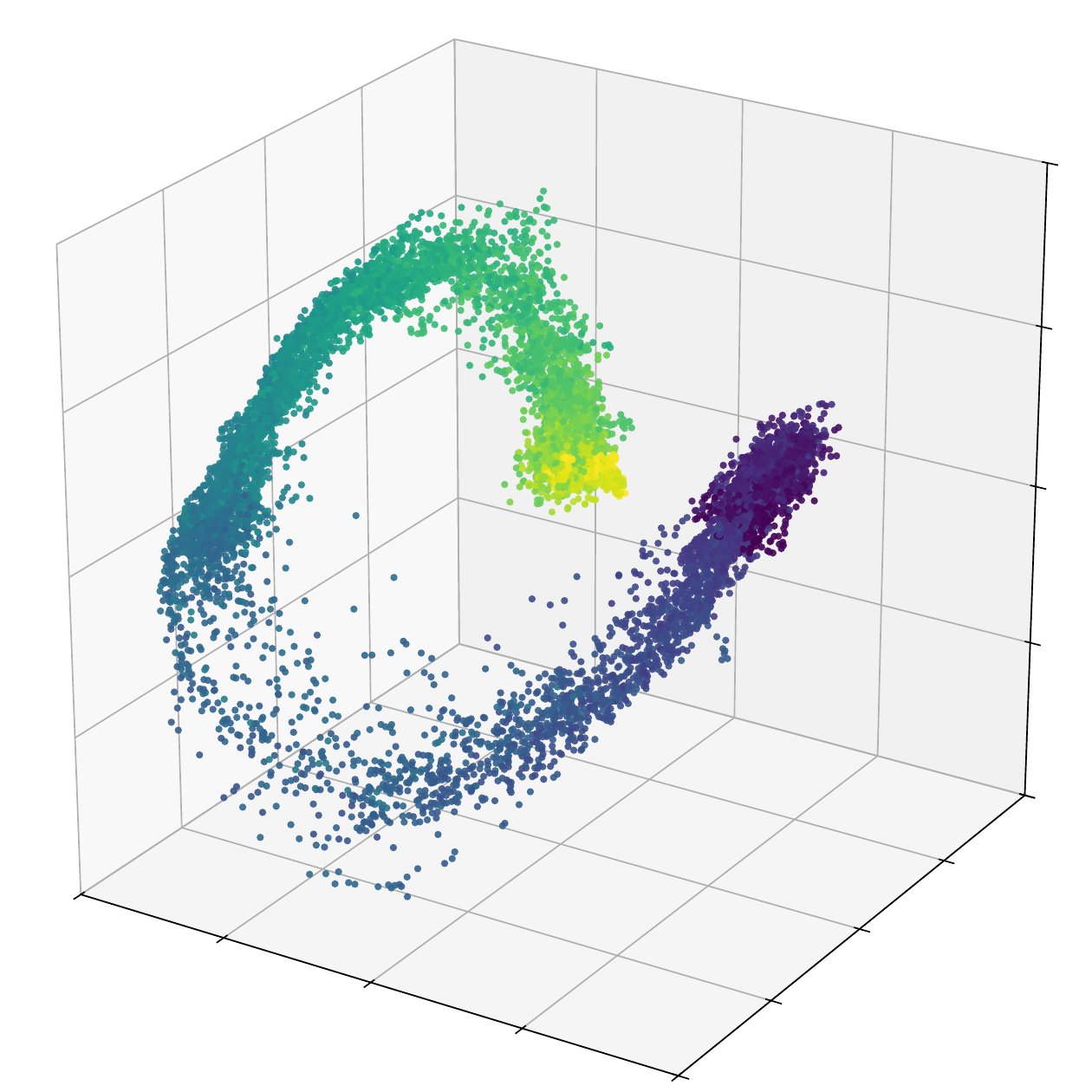}
        \caption{Progress Field}
        \label{fig:motivation_pf}
    \end{subfigure}

    \caption{
    Visualization of task progression and learned representations.
    The representation spaces are projected to three dimensions with PCA.
    (a)~End-effector trajectories of \textit{put both moka pots on the stove} task, with colors denoting normalized timesteps.
    (b)~Raw VLA representations.
    (c)~Representations from direct scalar progress prediction.
    (d)~Our Progress Field representations.
    }
    \label{fig:motivation}
\end{figure*}

To address the credit-assignment challenge, a natural solution is to model intermediate task progress and use it as dense feedback for policy improvement.
Prior works have explored various forms of progress-aware guidance for robot learning~\citep{ghasemipour2025self,zhai2025vision,liang2026robometer,yan2026progressvla,kim2026progvla}, demonstrating that denser intermediate feedback can substantially facilitate policy improvement under sparse task-level rewards.
Despite their architectural differences, task progress is commonly represented as an explicit scalar prediction.
While this formulation provides a simple and direct supervised learning target for capturing useful progress-related signals, it imposes limited structural constraints on how intermediate observations relate to the task goal, which may hinder effective transition-level credit assignment.
As illustrated in Figure~\ref{fig:motivation}, although task execution exhibits clear temporal progression in the physical workspace, raw VLA representations remain highly entangled across stages, while representations learned with a direct scalar prediction head still exhibit substantial overlap.
These observations motivate a more structured formulation of task progress over VLA representations, beyond direct scalar prediction.

Accordingly, we introduce \textbf{Progress Field Reinforcement Learning (PF-RL)}, which learns a structured goal-conditioned progress representation over pretrained VLA features and converts it into dense credit for policy optimization.
Concretely, a lightweight shared \textbf{Progress Field} head maps current and goal VLA representations into a compact progress space, where their geometric distance induces the corresponding goal-conditioned value.
The resulting goal-conditioned value geometry organizes observations according to their progress toward the task goal.
We learn this structure through complementary temporal and goal-structure objectives.
As illustrated in Figure~\ref{fig:motivation_pf}, the resulting representation exhibits a substantially clearer progress structure than both the raw VLA space and that learned with direct scalar prediction.
Based on the learned Progress Field, transition-level changes in goal-conditioned value yield progress-aware advantages, providing fine-grained credit for whether each transition advances toward or deviates from the task goal.
PF-RL then uses these advantages throughout offline-to-online reinforcement fine-tuning, providing dense progress-aware credit for both offline policy improvement and subsequent online adaptation. 
Experiments across simulated and real-world manipulation tasks demonstrate consistent improvements in policy performance and online adaptation across diverse settings.

\textbf{Our main contributions} are summarized as follows:
\begin{itemize}

    \item We introduce \textbf{Progress Field}, a lightweight structured progress representation learned over pretrained VLA features, which organizes current and goal observations in a shared goal-conditioned space where geometric distance induces the corresponding value.



    \item We propose \textbf{PF-RL}, a VLA reinforcement fine-tuning framework that converts transition-level changes in the learned Progress Field into dense progress advantages, providing fine-grained credit for both offline policy improvement and online reinforcement fine-tuning.
    

    \item 
    Extensive experiments on LIBERO, RoboTwin2.0, and real-world bimanual manipulation validate PF-RL and the benefits of its progress modeling over direct scalar prediction.

\end{itemize}
\section{Related Work}
\paragraph{Reinforcement Fine-Tuning~(RFT) for VLA} has recently emerged as an important post-training paradigm for VLA policies, enabling further improvement beyond imitation learning through interaction and task feedback.
Recent studies have adapted classical policy optimization and preference-based objectives to VLA fine-tuning, including PPO~\citep{lu2025vla}, GRPO~\citep{shao2024deepseekmath,liu2026can}, and DPO~\citep{rafailov2023direct, hung2025nora}, while open-source systems~\citep{li2026simplevla,yu2025rlinf} have facilitated large-scale VLA RFT and practical deployment.
Building on these foundations, recent methods explore specialized mechanisms, including RL optimization for flow-based VLAs~\citep {chen2025pirl}, self-referenced learning~\citep{fei2025srpo}, world-model-based policy optimization~\citep{zhu2025wmpo}, and latent reasoning~\citep{chen2026last}.
However, many VLA reinforcement fine-tuning methods still rely on sparse rollout-level outcomes, limiting credit assignment across intermediate transitions.
In contrast, PF-RL learns a structured goal-conditioned Progress Field over VLA representations and derives dense transition-level advantages for both offline and online policy optimization.

\paragraph{Progress Estimation} has been widely explored as a means of providing dense learning signals for robot policy learning.
Early representation-based approaches~\citep{ma2022vip,ma2023liv} pretrain goal-conditioned representations from large-scale human videos to derive dense rewards for downstream robot learning.
More recent embodied VLA and RL methods leverage large-scale robotic data to learn progress-aware signals, such as steps-to-go estimates and progress changes~\citep{ghasemipour2025self,zhai2025vision,tan2025robo,liang2026robometer}.
Building on learned progress estimators, \citet{yan2026progressvla} and \citet{bai2025evolve} use the resulting signals to guide downstream VLA policy improvement.
Meanwhile, \citet{kim2026progvla} estimate progress directly from internal VLA representations using auxiliary prediction heads.
Despite the architectural differences, many recent embodied approaches largely formulate progress as an explicit scalar prediction target, providing limited structural constraints on observation and goal relations and potentially restricting the utility of progress estimates for policy optimization.
In contrast, PF-RL learns a goal-conditioned value geometry over VLA representations and derives dense progress advantages from transition-level value changes to guide policy optimization.
\section{Preliminary}
\paragraph{Reinforcement Fine-Tuning of VLA Policies.} 

The reinforcement fine-tuning of VLA policies can be formulated as a language-conditioned Partially Observable Markov Decision Process~(POMDP)~\citep{cassandra1998survey}, characterized by
$\mathcal{M}=\langle\mathcal{S},\mathcal{A},P,R_\text{succ},\mathcal{I},\mathcal{O},\gamma,\rho_0\rangle$,
where \(\mathcal{S}\) represents the underlying environment and robot state space, \(\mathcal{A}\) represents the action space, \(\mathcal I\) refers to the space of natural language task descriptions, \(\mathcal O\) denotes the robot sensory observation space. \(P:\mathcal{S} \times \mathcal{A} \times \mathcal{S} \rightarrow \left[ 0, 1 \right]\) denotes the environment dynamics, \(R_\text{succ}: \mathcal S \times \mathcal I \rightarrow \{0,1\}\) is the binary reward function that detects whether the task is completed, \(\gamma \in (0, 1]\) is the discount factor, and \(\rho_0\) is the initial state distribution.
A VLA policy \(\pi\), parameterized as a multimodal policy \(\pi_\theta(\mathbf a_{t:t+H-1}|o_t, l)\), feeds on observation \(o_t\) of current timestep $t$ and language instruction $l\in\mathcal I$ to generate the action chunks \(\mathbf a_{t:t+H-1}\in \mathcal A^H\) for the next $H$ timesteps.
The RL objective for VLA fine-tuning is to maximize the expected return under the POMDP: \(\max_\theta \mathcal J(\theta)=\mathbb E_{\rho_0,P,\pi_\theta}{[\sum_{t=0}^\infty \gamma^t R_\text{succ}(s_t, l)]}\). 
Following policy-gradient methods~\citep{williams1992simple, silver2014deterministic}, the policy can be optimized using an advantage-weighted surrogate objective:
\begin{align}
  \mathcal L_\text{RFT}(\theta) = -\mathbb E_{\pi_\theta} \left[
    \sum_{t} A_{t}\log\pi_\theta(\mathbf a_{t:t+H-1}|o_t, l)
  \right],
  \label{eq:policy_gradient}
\end{align}
where $A_t$ is the advantage estimator. The advantage estimator can be further obtained from value function estimation~\citep{schulman2017proximal}, group-relative rollout comparison~\citep{shao2024deepseekmath}, or pairwise preference~\citep{rafailov2023direct}.

\begin{figure}[t] 
    \centering
      \includegraphics[width=1.0\textwidth]{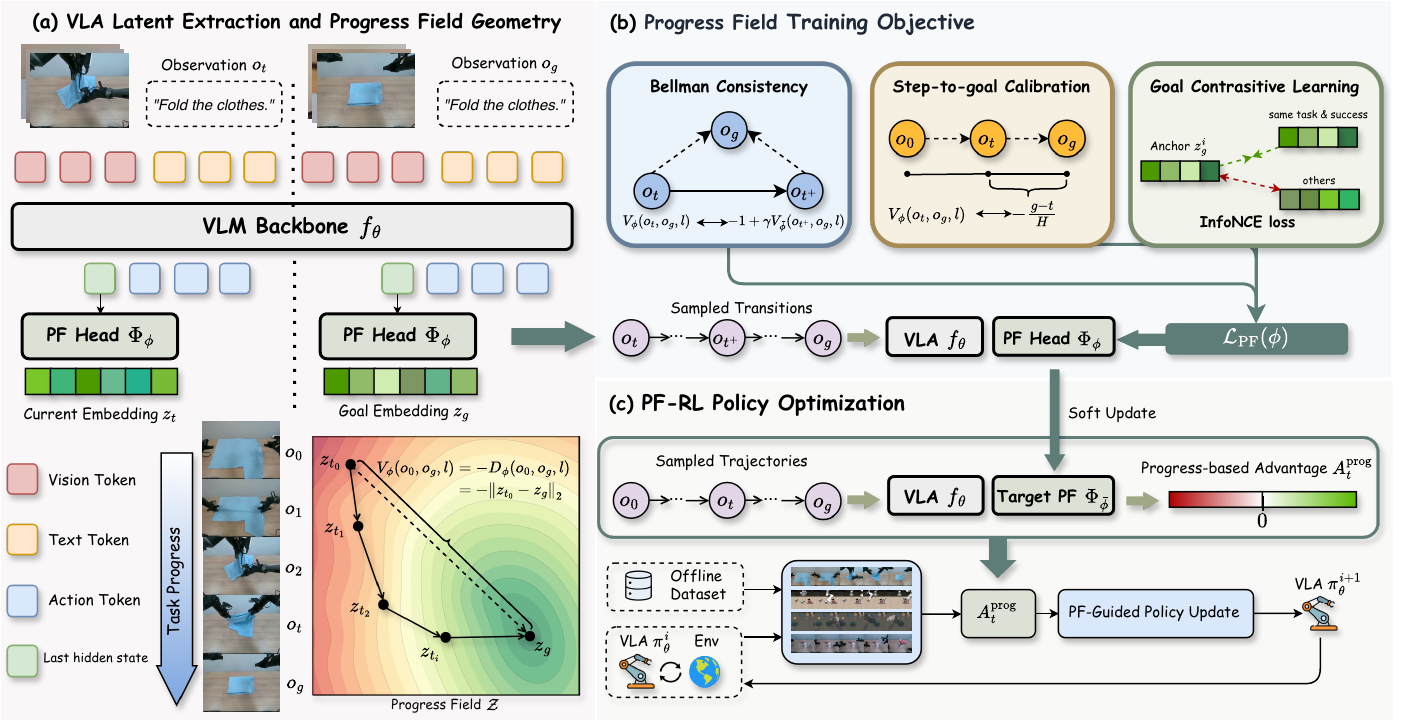} 
    \vspace{-20pt}
    \caption{An illustrative diagram of our proposed \textit{Progress Field Reinforcement Learning}.}
    \vspace{-10pt}
    \label{fig:framework}
\end{figure}

\section{Methodology}
\label{sec:method}
We introduce \textbf{Progress Field Reinforcement Learning (PF-RL)}, a VLA RFT framework that learns a structured goal-conditioned progress representation over pretrained VLA features and derives dense transition-level credit for policy optimization.
Our central idea is to model task progress through our proposed Progress Field rather than through direct scalar prediction.
For a fixed task goal, the Progress Field organizes observations in a shared progress space, where geometric distance to the goal induces the corresponding goal-conditioned value.
PF-RL consists of three key components: (i)~constructing a goal-conditioned Progress Field over VLA representations; (ii)~learning its geometry to reflect temporal progress and goal structure; and (iii)~deriving progress-aware advantages for offline and online policy optimization.
An overview of PF-RL is shown in Figure~\ref{fig:framework}.

\subsection{Progress Field as Goal-Conditioned Value Geometry}
\label{sec:progress_field}

\paragraph{VLA Latent Representation.}
Given the current observation $o_t$, goal observation $o_g$, and language instruction $l$, we extract their last-layer hidden representations from the VLA backbone:
\begin{align}
h_t &= f_\theta(o_t,l), \qquad
h_g = f_\theta(o_g,l),
\end{align}
where $f_\theta$ denotes the VLA backbone and $h_t$ and $h_g$ are the corresponding end-of-input hidden representations before action prediction.
We then map both representations into a compact progress space using a shared lightweight projection head $\Phi_\phi$ parameterized by $\phi$:
\begin{align}
z_t &= \Phi_\phi(h_t), \qquad
z_g = \Phi_\phi(h_g).
\label{eq:pf_embedding}
\end{align}

\paragraph{Goal-Conditioned Value Geometry.}
Given the current and goal embeddings $z_t$ and $z_g$, we define their distance and the corresponding goal-conditioned value as
\begin{align}
D_\phi(o_t,o_g,l)
=
\|z_t-z_g\|_2,
\qquad
V_\phi(o_t,o_g,l)
=
-D_\phi(o_t,o_g,l).
\label{eq:progress_value}
\end{align}
Following \citet{park2024foundation}, this parameterization assigns higher values to observations embedded closer to the task goal and lower values to those farther away.
For a fixed goal $o_g$ and instruction $l$, the resulting $V_\phi(\cdot,o_g,l)$ defines a goal-conditioned value geometry over observations, which we refer to as the \textbf{Progress Field}.
We refer to the projection head \(\Phi_\phi\) as the Progress Field head.
Task progress is therefore encoded through goal-relative geometry rather than predicted as a standalone scalar.
However, this geometric parameterization alone does not ensure alignment with temporal task progress along robot trajectories.
We next learn the Progress Field with objectives that impose temporal consistency and task-level goal structure.

\subsection{Progress Field Learning}
\label{sec:pf_learning}

Let $\mathcal D=\{\tau_i\}_{i=1}^{N}$ denote the trajectory data used for Progress Field learning, where each trajectory
$\tau_i=\{(o^i_t,\mathbf a^i_{t:t+H-1})\}_{t=0}^{T_i}$ is associated with a language instruction $l_i$, and $H$ denotes the action-chunk horizon.
The dataset is initialized with offline demonstrations and continuously augmented with newly collected trajectories during online fine-tuning.
During training, we construct tuples $(o_t,o_{t^+},o_u,o_g,l)$.
Here, $o_t$ is the current observation, and $o_{t^+}$ is the observation after executing one action chunk.
Since each chunk executes $H$ actions, $t^+=t+H$.
We take $o_g$ as the terminal observation of the same trajectory and randomly sample $o_u$ with $t^+<u<g$ as an auxiliary subgoal.
Both successful and failed trajectories are included in Progress Field training. For a failed trajectory, its terminal observation is treated as the observed endpoint of the trajectory rather than a task-completion goal.
Further details on goal assignment are provided in Appendix~\ref{app:exp_details}.
Based on these tuples, we learn the Progress Field using complementary objectives that align its geometric distance with temporal progress while preserving task-level goal structure.

\paragraph{Bellman consistency.}
The first objective imposes temporal consistency on the Progress Field.
Following~\citet{park2024foundation}, we use an expectile-weighted goal-conditioned TD objective:
\begin{align}
\mathcal L^g_{\mathrm{td}}(\phi)
=
\mathbb E_{(o_t,o_{t^+},o_g,l)}
\left[
\ell^\tau_2
\left(
-\mathbbm{1}(o_t\neq o_g)
+
\gamma V_{\bar\phi}(o_{t^+},o_g,l)
-
V_\phi(o_t,o_g,l)
\right)
\right],
\label{eq:pf_td}
\end{align}
where
$\ell^\tau_2(x)=|\tau-\mathbbm{1}(x<0)|x^2$
is the expectile regression loss and
$V_{\bar\phi}$ is computed using a slowly updated target projection with parameters $\bar\phi$.
Using \(\tau>0.5\) places greater weight on positive TD residuals, biasing the learned value toward higher-value outcomes rather than the conditional mean.
We set $\gamma=1$ in practice, consistent with the undiscounted steps-to-goal interpretation of the learned value.
This objective enforces temporal consistency of the goal-conditioned value, aligning its trajectory-wise changes with task progress.

\paragraph{Step-to-goal calibration.}
While Bellman bootstrapping enforces temporal consistency, we further calibrate Progress Field with the remaining temporal distance to the target.
For a target indexed by $g$, we define $\hat d_{t,g}=\frac{g-t}{H}$ to measure the remaining number of action-chunk intervals.
Since $V_\phi$ is parameterized as a negative distance, we optimize
\begin{align}
\mathcal L^g_{\mathrm{step}}(\phi)
=
\mathbb E_{(o_t,o_g,l)}
\left[
\left(
V_\phi(o_t,o_g,l)+\hat d_{t,g}
\right)^2
\right].
\label{eq:pf_step}
\end{align}
This objective calibrates the geometric distance with the remaining steps to the goal, providing an explicit temporal scale for the Progress Field.

\paragraph{Goal Contrastive Learning.}
While the temporal objectives structure progress within individual trajectories, they do not explicitly distinguish goal representations across different tasks.
We therefore apply an InfoNCE-style contrastive objective~\citep{oord2018representation} over terminal goal embeddings to capture task-level goal structure.
Let $e_g=\mathrm{Normalize}(z_g)$.
For anchor $i$, successful terminal observations from the same task form the positive set $\mathcal P(i)$, while terminal observations from failed trajectories of the same task and those from different tasks form the negative set $\mathcal N(i)$:
\begin{align}
\mathcal L_{\mathrm{con}}(\phi)
=
-\mathbb E_i
\left[
\log
\frac{
\sum_{j\in\mathcal P(i)}
\exp\left(
\mathrm{sim}(e_{g_i},e_{g_j})/\alpha
\right)
}{
\sum_{j\in\mathcal P(i)\cup\mathcal N(i)}
\exp\left(
\mathrm{sim}(e_{g_i},e_{g_j})/\alpha
\right)
}
\right],
\label{eq:pf_contrastive}
\end{align}
where $\mathrm{sim}(\cdot,\cdot)$ denotes cosine similarity and $\alpha$ is the contrastive temperature, encouraging same-task successful goals to cluster while separating failed and cross-task goals.

\paragraph{Auxiliary subgoal shaping.}
To better exploit the available geometric structure in long-horizon trajectories, we reuse later observations as auxiliary subgoals. 
For the auxiliary subgoal $o_u$, we apply the Bellman-consistency and step-to-target objectives, while the terminal-goal branch additionally includes goal contrastive learning. The overall Progress Field objective is
\begin{align}
\mathcal L_{\mathrm{PF}}(\phi)
=
\underbrace{
\mathcal L^g_{\mathrm{td}}
+
\mathcal L^g_{\mathrm{step}}
+
\mathcal L_{\mathrm{con}}
}_{\mathcal L_{\mathrm{goal}}}
+
\lambda_{\mathrm{sub}}
\underbrace{
\left(
\mathcal L^u_{\mathrm{td}}
+
\mathcal L^u_{\mathrm{step}}
\right)
}_{\mathcal L_{\mathrm{sub}}}.
\label{eq:pf_loss}
\vspace{-0.5cm}
\end{align}
After each update, the target parameters $\bar\phi$ are softly updated from $\phi$ to stabilize TD bootstrapping. Meanwhile, gradients from $\mathcal L_{\mathrm{PF}}$ are stopped at the VLA representation $h$, so the progress-aware geometry is learned through the Progress Field head $\Phi_\phi$ while preserving the underlying VLA representation for further policy optimization.

\subsection{PF-RL Policy Optimization}
\label{sec:pfrl_po}

Based on the learned Progress Field, we define a progress-aware advantage from transition-level changes in the goal-conditioned value.
Specifically, for a transition $o_t\rightarrow o_{t^+}$, we compute
\begin{align}
A^{\mathrm{prog}}_t
=
-\mathbbm{1}(o_t\neq o_g)
+
\gamma V_{\bar\phi}(o_{t^+},o_g,l)
-
V_{\bar\phi}(o_t,o_g,l).
\label{eq:progress_advantage}
\end{align}
This advantage provides fine-grained credit by evaluating progress of each transition toward the goal relative to the learned temporal scale.
For successful trajectories, the terminal observation naturally serves as the goal reference. 
For failed trajectories, the terminal observation does not represent task completion.
During online fine-tuning, we therefore replace its goal embedding with a per-task goal prototype only when computing $A^{\mathrm{prog}}_t$.
The prototype is maintained through an exponential moving average of successful goal embeddings from the same task.
This provides a success-based reference for evaluating progress on failed online trajectories.
PF-RL then uses $A^{\mathrm{prog}}_t$ for offline policy improvement and subsequent online reinforcement fine-tuning, as described below.

\paragraph{Offline Optimization.} 
In the offline stage, PF-RL reweights the native VLA imitation objective using $A_t^{\mathrm{prog}}$ following the principle of advantage-weighted regression~\citep{peng2019advantage}:
\begin{align}
\mathcal L_{\mathrm{off}}(\theta)
=
\mathbb E_{(o_t,\mathbf a_{t:t+H-1},l)\sim\mathcal D}
\left[
\exp\left(\frac{A^{\mathrm{prog}}_t}{\beta}\right)
\ell_{\mathrm{VLA}}
\left(\theta; o_t,l,\mathbf a_{t:t+H-1}\right)
\right],
\label{eq:offline_awbc}
\end{align}
where $\beta$ controls the sharpness of advantage weighting and $\ell_{\mathrm{VLA}}$ denotes the backbone-specific imitation loss.
In simulation, $\ell_{\mathrm{VLA}}$ is the token-level cross-entropy loss of OpenVLA-OFT, whereas in real-world experiments, it is the flow-matching objective of $\pi_{0.5}$.
This weighting favors transitions with higher progress advantages.

\paragraph{Online Fine-Tuning.}
In simulation, sparse terminal rewards provide rollout-level task feedback, while $A^{\mathrm{prog}}_t$ provides dense progress-aware advantages for intermediate transitions.
For a group of $G$ rollouts with trajectory-level task rewards
$\{R_i\}_{i=1}^{G}$, we compute the GRPO~\citep{shao2024deepseekmath} advantage by group-wise normalization and combine it with the transition-level progress advantage:
\begin{align}
A^{\mathrm{GRPO}}_i
&=
\frac{
R_i-\operatorname{mean}(\{R_j\}_{j=1}^{G})
}{
\operatorname{std}(\{R_j\}_{j=1}^{G})+\varepsilon
},
\qquad
A^{\mathrm{sim}}_{i,t}
=
A^{\mathrm{GRPO}}_i
+
\lambda_{\mathrm{prog}}
A^{\mathrm{prog}}_{i,t},
\label{eq:sim_advantage}
\end{align}
where $\lambda_{\mathrm{prog}}$ controls the contribution of the progress advantage. 
We then optimize the policy with the clipped GRPO objective.
Defining the policy ratio as
\(
r_{i,t}(\theta)
=
\frac{
\pi_\theta(
\mathbf a_{t:t+H-1}\mid o_t,l
)
}{
\pi_{\theta_{\mathrm{old}}}(
\mathbf a_{t:t+H-1}\mid o_t,l
)}
\), we optimize
\begin{align}
\mathcal L_{\mathrm{GRPO}}(\theta)
=
-
\mathbb E
\left[
\min
\left(
r_{i,t}(\theta)A^{\mathrm{sim}}_{i,t},\,
\mathrm{clip}
\left(
r_{i,t}(\theta),1-\epsilon,1+\epsilon
\right)
A^{\mathrm{sim}}_{i,t}
\right)
\right].
\label{eq:grpo}
\end{align}

For real-robot fine-tuning, where interaction is costly and online data are limited, we use $A^{\mathrm{prog}}_t$ to reweight the flow-matching objective of $\pi_{0.5}$ according to Eq.~\ref{eq:offline_awbc}.
This conservative update emphasizes transitions with higher progress advantages while retaining the native policy objective.

Notably, Progress Field learning is interleaved with both offline and online policy optimization rather than performed as a separate pre-training stage.
This keeps the learned geometry synchronized with evolving VLA representations while allowing newly collected trajectories to continuously refine the Progress Field.
Overall, PF-RL leverages the learned Progress Field to provide fine-grained credit throughout VLA policy optimization.
Detailed pseudocode is provided in Appendix~\ref{app:pseudocode}.

\section{Experiments}
\label{sec:exp}
We evaluate PF-RL on simulated and real-world robotic manipulation tasks. 
Our experiments aim to answer the following questions: 
(1)~Can PF-RL effectively improve VLA reinforcement fine-tuning across simulation and real-world settings?~(Section~\ref{sec:exp-sim} and Section~\ref{sec:exp-real}) 
(2)~Does the Progress Field formulation provide more effective progress modeling than direct scalar prediction?~(Section~\ref{sec:exp_ablation} and Appendix~\ref{app:progress_modeling})
(3)~How do the key design choices of PF-RL affect policy performance?~(Section~\ref{sec:exp_ablation} and  Appendix~\ref{app:additional_ablation}) 
(4)~Does the learned Progress Field capture structured goal-conditioned value geometry that reflects task progress?~(Section~\ref{sec:exp_vis}, Appendix~\ref{app:goal_embedding_vis}, and Appendix~\ref{app:pf_visualizations})

\subsection{Simulation Experiments.}
\label{sec:exp-sim}
\paragraph{Experimental Setup.}
We evaluate PF-RL on two simulated manipulation benchmarks, LIBERO and RoboTwin2.0. 
LIBERO~\citep{liu2023libero} comprises the Spatial, Object, Goal, and Long suites, with 10 tasks per suite and 40 tasks in total. 
For each task, we sample 30 expert demonstrations from the official dataset for offline training. 
RoboTwin2.0~\citep{chen2025robotwin} features bimanual manipulation tasks with complex and contact-rich interactions. 
We construct the offline training data following the protocol in~\citet{li2026simplevla}.
Compared with LIBERO, where recent VLA methods already achieve strong performance, RoboTwin2.0 offers a more challenging benchmark for evaluating policy improvement under complex manipulation dynamics. 
For all simulation experiments, we initialize PF-RL from the OpenVLA-OFT checkpoint provided by~\citet{li2026simplevla}.
We use RLinf-GRPO as our primary controlled RFT baseline on LIBERO. Both methods start from the same OpenVLA-OFT checkpoint and are trained on the same offline data for the same number of optimization steps, with PF-RL applying its offline objective and RLinf-GRPO using standard SFT. 
The resulting policies are then fine-tuned online using their respective objectives under the same training budget.
We additionally report published results of SFT, RFT, and progress-aware VLA methods for reference.
More details are provided in Appendix~\ref{app:sim_details}.

\begin{figure}[t]
    \centering
    \captionof{table}{
    Success Rate  \((\%)\) comparison on LIBERO~\cite{liu2023libero}.
    The best results are in \textbf{bold}, while \(\pm\) denotes standard deviation over 3 independent training seeds. 
    $\dagger$ denotes baselines reproduced under the same offline-to-online
    budgets; other baseline results are from the original papers.
    }
    \label{tab:libero_results}
    \begin{tabularx}{\columnwidth}{>{\raggedright\arraybackslash}X|cccc|c}
    \toprule
    \textbf{Models}
        & \textbf{Spatial}
        & \textbf{Object}
        & \textbf{Goal}
        & \textbf{Long}
        & \textbf{Average} \\
    \midrule
    \multicolumn{6}{l}{\textcolor{gray}{\textit{SFT baselines}}} \\
    $\pi_0$~\citep{black2024pi_0}
        & 96.8 & 98.8 & 95.8 & 85.2 & 94.2 \\
    $\pi_{0.5}$~\citep{intelligence2025pi_}
        & 98.8 & 98.2 & 98.0 & 92.4 & 96.9 \\
    OpenVLA~\citep{kim2024openvla}
        & 84.7 & 88.4 & 79.2 & 53.7 & 76.5 \\
    \mbox{OpenVLA-OFT~\citep{kim2025fine}}
        & 97.6 & 98.4 & 97.9 & 94.5 & 97.1 \\
    \midrule
    \multicolumn{6}{l}{\textcolor{gray}{\textit{RFT baselines}}} \\
    GRAPE~\citep{zhang2024grape}
        & 88.5 & 92.1 & 83.1 & 57.2 & 80.2 \\
    TGRPO~\citep{chen2025tgrpo}
        & 90.4 & 92.2 & 81.0 & 59.2 & 80.7 \\
    VLA-RL~\citep{lu2025vla}
        & 90.2 & 91.8 & 82.2 & 59.8 & 81.0 \\
    RLinf-GRPO$^\dagger$~\citep{yu2025rlinf}
        & 97.6\scalebox{0.6}{$\pm 0.4$} 
        & 98.7\scalebox{0.6}{$\pm 0.6$} 
        & 95.2\scalebox{0.6}{$\pm 0.7$} 
        & 96.8\scalebox{0.6}{$\pm 0.5$} 
        & 97.1 \\
    \midrule

    \multicolumn{6}{l}
        {\textcolor{gray}{\textit{Progress-aware baselines}}} \\
    ProgressVLA~\citep{yan2026progressvla}
        & 88.2 & 96.4 & 87.2 & 66.2 & 84.5 \\
    ProgVLA~\citep{kim2026progvla}
        & 87.6 & 96.0 & 92.0 & 88.6 & 91.1 \\
    EVOLVE-VLA~\citep{bai2025evolve}
        & 95.4 & 97.4 & 95.8 & 94.4 & 95.8 \\
    \midrule

\multicolumn{6}{l}{\textit{OpenVLA-OFT + PF-RL}} \\

PF-RL (offline)
    & 87.4\scalebox{0.6}{$\pm 0.8$} 
    & 77.8\scalebox{0.6}{$\pm 0.5$} 
    & 84.6\scalebox{0.6}{$\pm 1.2$} 
    & 76.8\scalebox{0.6}{$\pm 1.4$} & 81.7 \\

\rowcolor{gray!15}
+ online
    & \textbf{99.2}\scalebox{0.6}{$\pm 0.7$}
    & \textbf{99.5}\scalebox{0.6}{$\pm 0.3$} 
    & \textbf{98.8}\scalebox{0.6}{$\pm 1.0$}
    & \textbf{99.0}\scalebox{0.6}{$\pm 0.5$}
    & \textbf{99.1} \\

    \bottomrule

    \end{tabularx}
    
    \vspace{4pt}
    \includegraphics[width=\columnwidth]
    {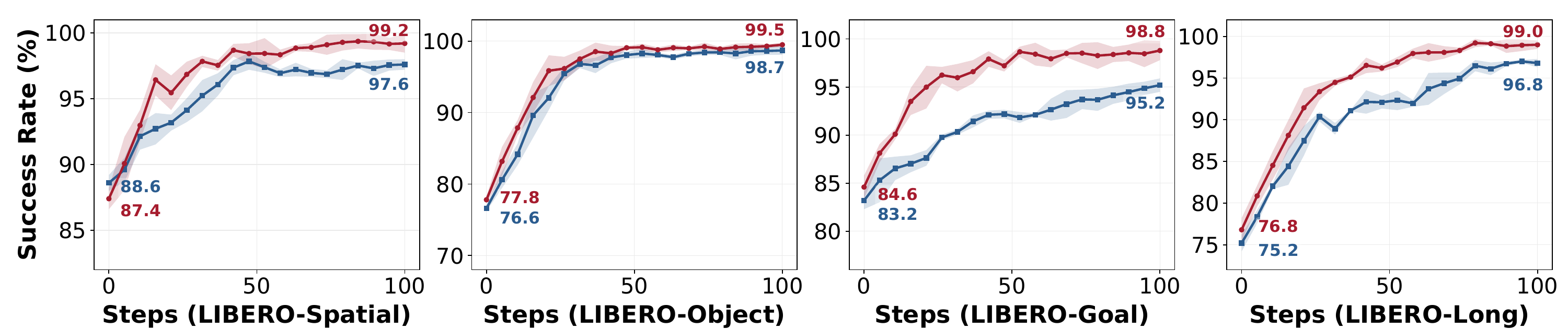}

    \captionof{figure}{Online fine-tuning curves of \textcolor[HTML]{A61C2E}{\textbf{PF-RL}} and \textcolor[HTML]{2B5C8F}{\textbf{RLinf-GRPO}} across the LIBERO task suites.}
    \label{fig:learning_curves}

    \vspace{-13pt}
\end{figure}

\paragraph{Results Analysis.}
Table~\ref{tab:libero_results} shows that PF-RL achieves an average success rate of 99.1\% on LIBERO after online fine-tuning. 
Compared with RLinf-GRPO under matched offline-to-online training budgets, PF-RL improves performance across all four suites, with gains of 2.0 percentage points on average and 2.2 percentage points on LIBERO-Long. 
The improvement on LIBERO-Long suggests that dense progress feedback can facilitate policy learning in temporally extended manipulation tasks.
Figure~\ref{fig:learning_curves} further shows faster convergence and higher final success rates, supporting the effectiveness of progress-aware feedback for policy optimization. 
PF-RL also achieves a higher reported average success rate than the progress-aware baselines. Meanwhile, Table~\ref{tab:robotwin2} shows consistent improvements over RLinf-GRPO across all four evaluated tasks on RoboTwin2.0, providing further evidence of its effectiveness in challenging bimanual manipulation.

\subsection{Real-World Experiments}
\label{sec:exp-real}
\paragraph{Experimental Setup.} 
To evaluate PF-RL in real-world settings, we conduct experiments on the Agilex Cobot Magic dual-arm platform. 
We consider four challenging manipulation tasks: \emph{Fold Clothes}, \emph{Stack all Plates}, \emph{Put all Blocks into Bucket}, and \emph{Fill Vase and Arrange Flowers}. We train \emph{Fold Clothes} separately in a single-task setting, while the latter three tasks are jointly trained in a multi-task setting.
For all real-world experiments, we initialize PF-RL from the official JAX checkpoint of $\pi_{0.5}$~\citep{intelligence2025pi_}, and then perform two iterations of real-world reinforcement fine-tuning following \citet{intelligence2025pi06vlalearnsexperience}, with 20 rollouts collected per task in each iteration.
To improve the safety and efficiency of real-world exploration, we adopt human-in-the-loop intervention~\citep{luo2025precise}, allowing corrective intervention during rollout collection. 
More detailed settings are provided in Appendix~\ref{app:real_details}.
\begin{table*}[t]
    \centering
    \small
    \setlength{\tabcolsep}{5pt}
    \caption{Success Rate \((\%)\) comparison on RoboTwin2.0~\cite{chen2025robotwin}.
    }
    \vspace{-7pt}
    \label{tab:robotwin2}

    \begin{tabular}{l|ccccc}
        \toprule
        Method
        & \makecell{Beat Block\\Hammer}
        & \makecell{Pick Dual\\Bottles}
        & \makecell{Stack Bowls\\Two}
        & \makecell{Place Empty\\Cup}
        & Avg. \\
        \midrule
        $\pi_0$ (SFT)~\citep{black2024pi_0}
            & 59 & 50 & 53 & 60 & 56 \\
        RDT (SFT)~\citep{liu2024rdt}
            & 22 & 18 & 42 & 42 & 31 \\
        OpenVLA-OFT~\citep{kim2025fine}
            & 13 & 11 & 82 & 81 & 47 \\
        RLinf-GRPO$^\dagger$~\citep{yu2025rlinf}
            & 82 & 70 & 71 & 94 & 79 \\
        \midrule
        \rowcolor{gray!15} OpenVLA-OFT+PF-RL~(offline\(\rightarrow\)online)
                & {\small 25$\rightarrow$\textbf{91}}
                & {\small 19$\rightarrow$\textbf{93}}
                & {\small 53$\rightarrow$\textbf{86}}
                & {\small 64$\rightarrow$\textbf{96}}
                & \textbf{92} \\
                
        \bottomrule
    \end{tabular}
\end{table*}

\begin{table*}[t]
\centering
\newcommand{\graygap}[1]{\textsubscript{\textcolor{gray!80}{\tiny(#1)}}} 

\caption{
Real-world comparison under the original setting and three visual shifts: unseen objects (-Obj), background (-BG), and lighting (-L). All baselines are reproduced under the same protocol and evaluated over 20 rollouts per setting. 
Detailed settings are provided in Appendix~\ref{app:real_world_tasks}.
}
\label{tab:real_world_results}

    \vspace{-7pt}
\setlength{\tabcolsep}{3.5pt} 
\small
\begin{tabularx}{\textwidth}{l | *{4}{>{\centering\arraybackslash}X} | *{4}{>{\centering\arraybackslash}X}}
\toprule
\multirow{2}{*}{\textbf{Methods}} & \multicolumn{4}{c|}{\textbf{Fold Clothes}} & \multicolumn{4}{c}{\textbf{Stack all Plates}} \\
\cmidrule(lr){2-5} \cmidrule(lr){6-9}
& Original & -Obj & -BG & -L & Original & -Obj & -BG & -L \\
\midrule
$\pi_{0.5}$~\citep{intelligence2025pi_}  & 75 & 65\graygap{-10\%} & 70\graygap{-5\%} & 55\graygap{-20\%} & 70 & 65\graygap{-5\%} & 55\graygap{-15\%} & 65\graygap{-5\%} \\
 Evo-RL~\citep{evorl2026}     & 90 & 70\graygap{-20\%} & 80\graygap{-10\%} & 65\graygap{-25\%} & 90 & 85\graygap{-5\%} & 80\graygap{-10\%} & 85\graygap{-5\%} \\\midrule
\rowcolor{gray!15} $\pi_{0.5}$+PF-RL~(offline\(\rightarrow\)online)     & 80\(\rightarrow\)\textbf{95} & 90\graygap{-5\%}  & 90\graygap{-5\%}  & 85\graygap{-10\%} & 85\(\rightarrow\)\textbf{95} & 95\graygap{-0\%} & 90\graygap{-5\%} & 90\graygap{-5\%} \\
\midrule
\midrule

\multirow{2}{*}{\textbf{Methods}} & \multicolumn{4}{c|}{\textbf{Put all Blocks into Bucket}} & \multicolumn{4}{c}{\textbf{Fill Vase and Arrange Flowers}} \\
\cmidrule(lr){2-5} \cmidrule(lr){6-9}
& Original & -Obj & -BG & -L & Original & -Obj & -BG & -L \\
\midrule
$\pi_{0.5}$~\citep{intelligence2025pi_} & 70 & 55\graygap{-15\%} & 60\graygap{-10\%} & 35\graygap{-35\%} & 50 & 40\graygap{-10\%} & 35\graygap{-15\%} & 30\graygap{-20\%} \\
Evo-RL~\citep{evorl2026}      & 85 & 80\graygap{-5\%} & 75\graygap{-10\%} & 65\graygap{-20\%} & 75 & 60\graygap{-15\%} & 55\graygap{-20\%} & 65\graygap{-10\%} \\\midrule
\rowcolor{gray!15} $\pi_{0.5}$+PF-RL~(offline\(\rightarrow\)online)      & 80\(\rightarrow\)\textbf{90} & 80\graygap{-10\%}  & 85\graygap{-5\%}  & 75\graygap{-15\%} & 65\(\rightarrow\)\textbf{85} & 75\graygap{-10\%} & 75\graygap{-10\%} & 80\graygap{-5\%} \\
\bottomrule
\end{tabularx}
\end{table*}

\paragraph{Results Analysis.}
As shown in Table~\ref{tab:real_world_results}, PF-RL consistently improves policy performance across all evaluated real-world tasks. 
Under the Original setting, offline PF-RL already improves upon the $\pi_{0.5}$ policy, while further adaptation with newly collected real-robot trajectories yields substantial additional gains and outperforms online Evo-RL. 
Meanwhile, the resulting PF-RL policy also maintains strong performance under the evaluated unseen-object, background, and lighting shifts.
Overall, these results demonstrate the effectiveness of progress-aware feedback for both offline policy improvement and real-world online adaptation. 

\begin{figure}[htbp]
    \centering
    \captionsetup[subfigure]{skip=2pt}
    \begin{subfigure}[t]{0.32\linewidth}
        \centering
        \includegraphics[width=\linewidth]{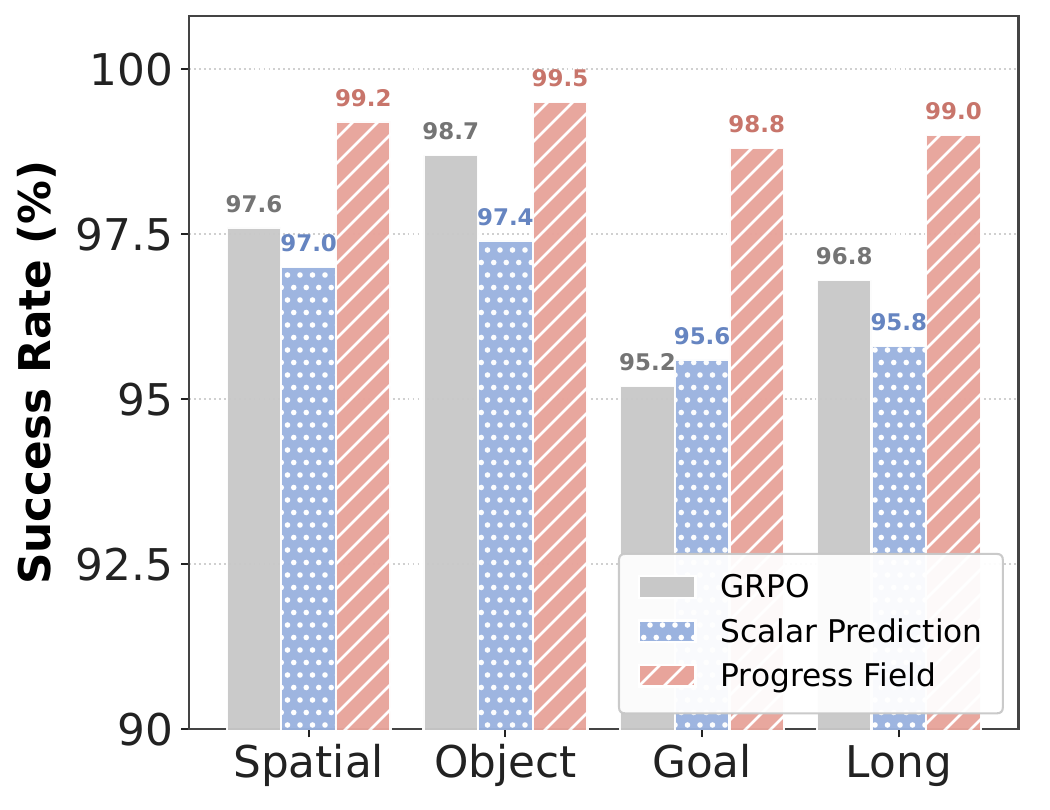}
        \caption{LIBERO Final Performance}
        \label{fig:progress_formulation_libero}
    \end{subfigure}
    \hfill
    \begin{subfigure}[t]{0.32\linewidth}
        \centering
        \includegraphics[width=\linewidth]{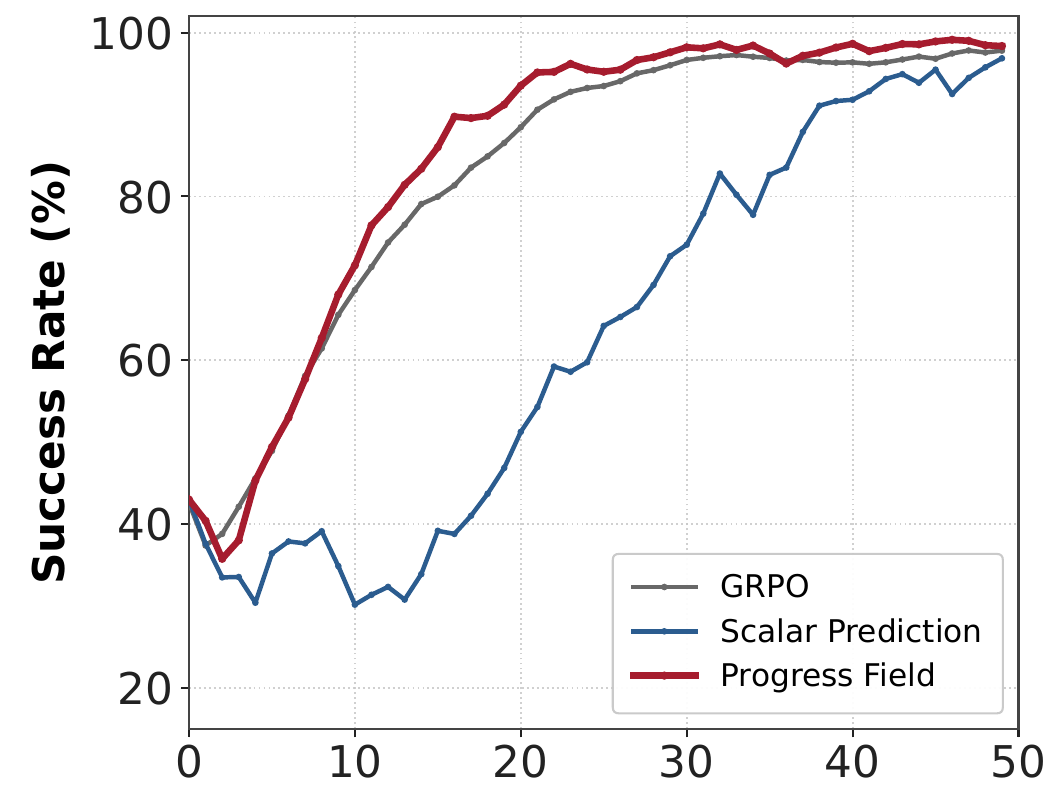}
        \caption{LIBERO-Goal Task5 }
        \label{fig:progress_transfer_goal}
    \end{subfigure}
    \hfill
    \begin{subfigure}[t]{0.32\linewidth}
        \centering
        \includegraphics[width=\linewidth]{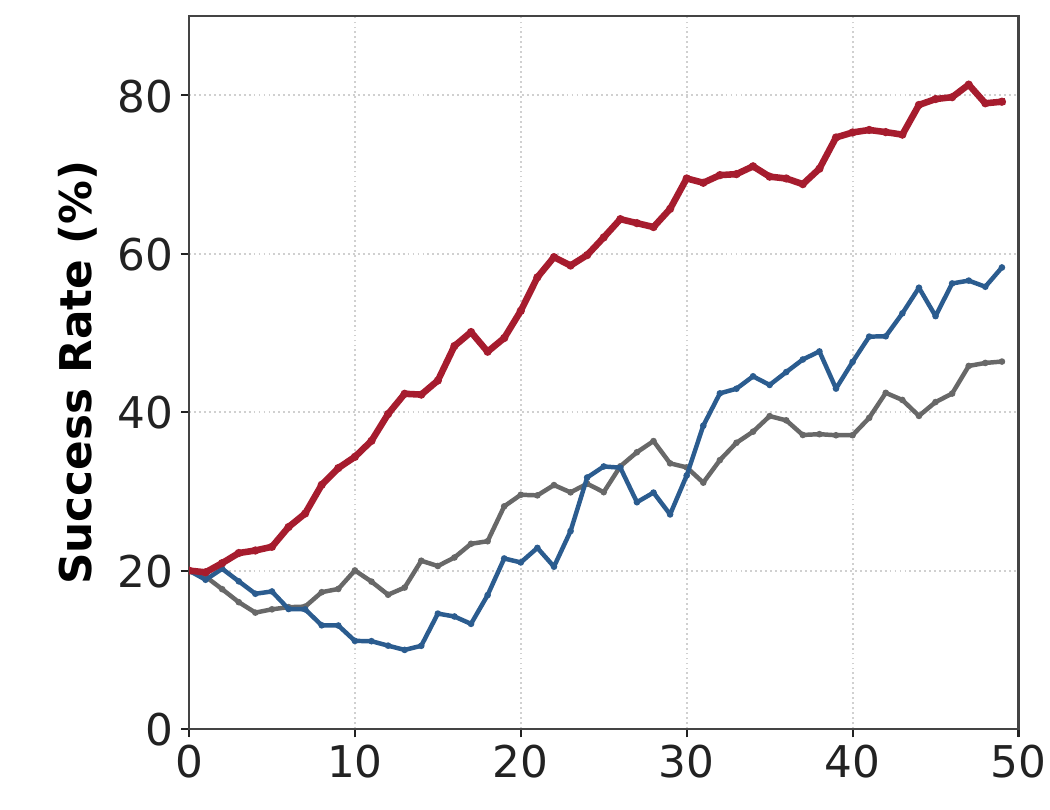}
        \caption{LIBERO-Long Task0 }
        \label{fig:progress_transfer_long}
    \end{subfigure}
    \vspace{-8pt}
\caption{
    Scalar prediction vs. Progress Field. (a)~Final performance on four LIBERO suites. (b--c)~Online adaptation on held-out tasks, with progress models trained on the remaining nine tasks.
}
    \label{fig:progress_modeling_analysis}
\end{figure}

\subsection{Ablation Study}

\paragraph{Effect of the Progress Field Formulation.}
Figure~\ref{fig:progress_formulation_libero} shows that Progress Field outperforms direct scalar prediction across all four LIBERO suites under the same training pipeline.
Table~\ref{tab:scalar_prediction_variants} further shows that this advantage persists when both methods omit contrastive learning, and that Progress Field outperforms the MLP readout with the same projection architecture and training objectives.
To assess cross-task transfer, we train the progress models offline on the remaining nine tasks for each held-out task.
The results demonstrate that Progress Field enables stronger online adaptation on held-out tasks, suggesting better cross-task transfer.
Further analysis is provided in Appendix~\ref{app:progress_modeling}.

\label{sec:exp_ablation}
\begin{wrapfigure}{r}{0.44\textwidth}
    \vspace{-1pt} 
    \centering
    \includegraphics[width=\linewidth]{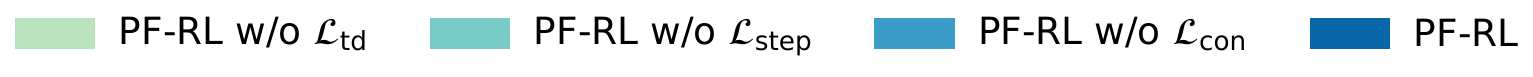}\\
    \includegraphics[width=\linewidth]{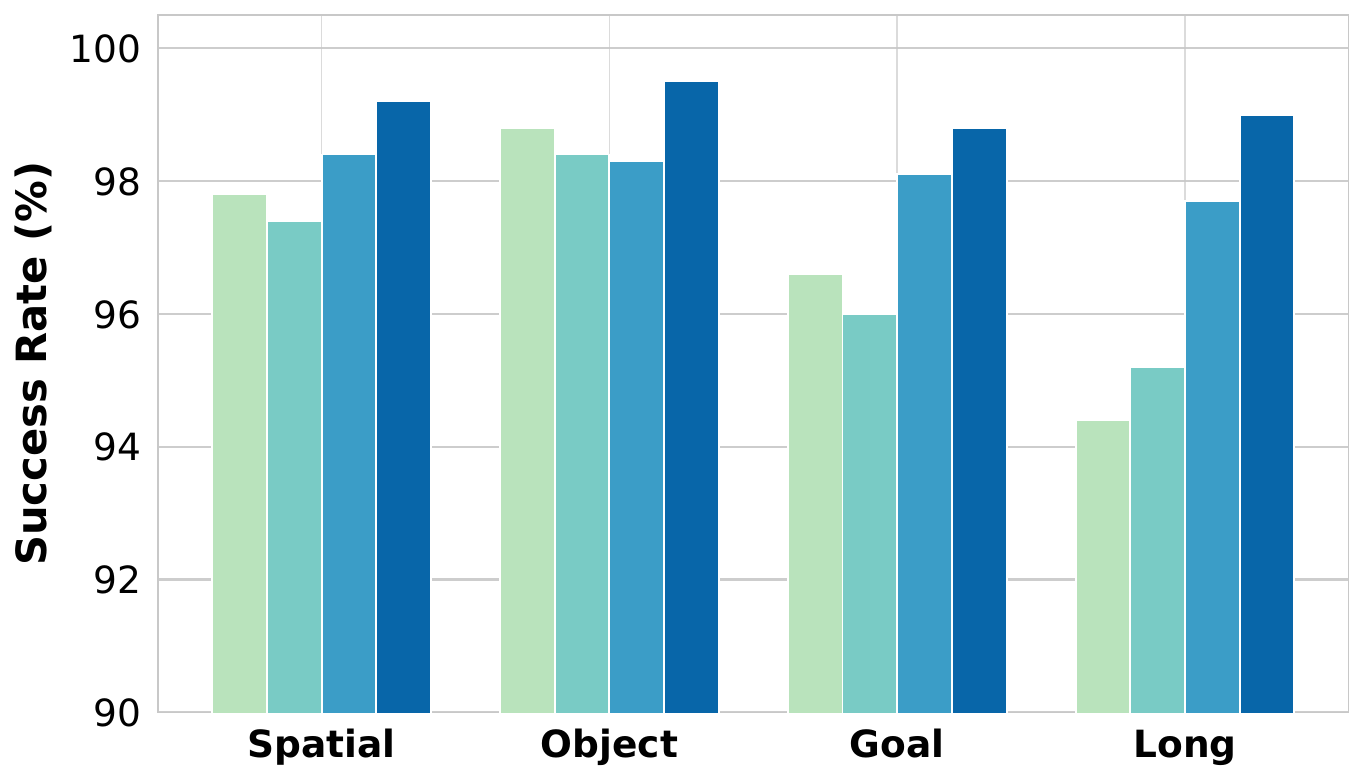}
    \vspace{-17pt}
    \caption{Ablation of Progress Field training objectives on the four LIBERO suites.}
    \label{fig:tpae_loss_ablation}
    \vspace{-5pt} 
\end{wrapfigure}
\paragraph{Effect of Progress Field Training Objectives.}
As shown in Figure~\ref{fig:tpae_loss_ablation}, the complete Progress Field training objective consistently achieves the strongest performance across all four LIBERO suites, with detailed numerical results reported in Table~\ref{tab:pf_objective_ablation}.
Removing the Bellman consistency loss causes the most pronounced degradation on LIBERO-Long, highlighting its importance for propagating goal-conditioned values over long-horizon trajectories.
Meanwhile, removing the step-to-goal objective leads to broad performance drops by weakening the temporal calibration of the learned geometry, while removing the goal-contrastive objective mainly reduces task-level goal discrimination.
Moreover, the HILP-inspired TD-only metric variant as shown in Table~\ref{tab:pf_objective_ablation} also underperforms the full objective.
Together, these results show that the three objectives provide complementary temporal, metric, and task-level constraints for learning a well-structured goal-conditioned value geometry.
Detailed scores are reported in Appendix~\ref{app:additional_ablation_objective}.

\begin{figure*}[t]
    \includegraphics[width=\linewidth]{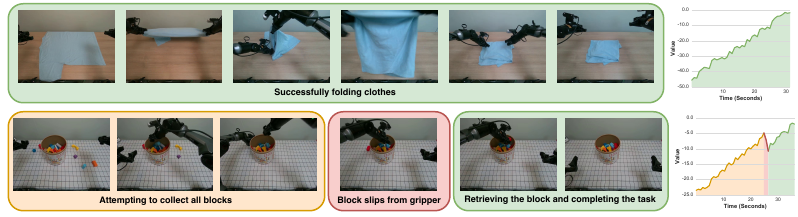}
    \vspace{-10pt}
    \caption{
    Qualitative visualization of goal-conditioned values induced by the learned Progress Field during real-world task execution.
    The top shows successful cloth folding, while the bottom illustrates putting all blocks into bucket with an intermediate grasp failure followed by successful recovery. 
    More visualizations are provided in Appendix~\ref{app:pf_visualizations}.}
    \label{fig:value_vis}
    \vspace{-13pt}
\end{figure*}
\subsection{Visualization}
\label{sec:exp_vis}

Figure~\ref{fig:value_vis} further visualizes the goal-conditioned value induced by the learned Progress Field along representative real-world executions.
During successful cloth folding, the estimated value exhibits a clear overall increase as the manipulation progresses toward the goal.
More notably, in the \emph{Put all Blocks into Bucket} task, the value first increases with task progress, drops sharply when the red block slips from the gripper, and rises again after the robot recovers and successfully completes the task.
Such non-monotonic changes indicate that the learned value responds to current events rather than simply increasing with elapsed time.
These observations show that the learned goal-conditioned value responds consistently to task advancement, execution degradation, and subsequent recovery, supporting the effectiveness of the Progress Field for progress-aware advantage estimation.


\section{Conclusion}

In this work, we study the credit-assignment challenge in VLA reinforcement fine-tuning, where sparse task-level outcomes provide limited guidance for intermediate transitions.
To address this problem, we propose PF-RL, which learns a lightweight structured progress representation over pretrained VLA features and derives dense progress advantages from the resulting goal-conditioned value geometry for both offline and online policy optimization.
Extensive experiments on LIBERO, RoboTwin2.0, and real-world bimanual manipulation tasks demonstrate consistent policy improvements, while controlled comparisons and representation analyses further support the benefits of Progress Field over direct scalar prediction.
Overall, our results highlight goal-conditioned value geometry as an effective formulation for learning progress-aware credit in VLA RFT.

\subsection*{AI use statement}
In this work, we used generative AI tools to help refine the conceptual framework of PF-RL, 
articulate and refine the research idea initially developed by the authors,
provide feedback on research methodology and experimental design, assist with method implementation and translation, and support the analysis and discussion of experimental results. 
We did not use generative AI tools to generate synthetic datasets, formulate mathematical claims, provide critical ingredients for proving mathematical claims, assist in the writing of proofs, or clean and reformat datasets. Qualitative and thematic data analysis was not applicable to this work.
We have reviewed all AI-assisted work, including carefully assessing suggestions on research methodology and experimental design, verifying that the AI-assisted analysis of experimental results was supported by the underlying data, and inspecting AI-assisted code for correctness and consistency with the described method.
We take full responsibility for all AI-assisted content in this work.

\subsection*{Reproducibility statement}
We provide comprehensive details to facilitate the reproduction of our results.
The complete formulation of PF-RL and its training procedures are presented in Section~\ref{sec:method} and Appendix~\ref{app:pseudocode}, including the
construction and sampling of training tuples, the Progress Field training
objective, and the offline and online policy optimization pipelines.
All experimental settings, model hyperparameters, and hardware specifications
are reported in Appendix~\ref{app:exp_details}.
Our simulation implementation builds on \href{https://github.com/RLinf/RLinf}{RLinf},
while our real-robot implementation uses \href{https://github.com/Physical-Intelligence/openpi}{OpenPI} as the underlying VLA codebase.
For human-in-the-loop data collection during real-world online RL, we build on the open-source \href{https://github.com/MINT-SJTU/Evo-RL}{Evo-RL}.
Together, the mathematical formulations, implementation details, and experimental configurations provided in the paper and appendices specify the complete experimental protocol of this work.

\bibliography{iclr2027_conference}
\bibliographystyle{iclr2027_conference}

\appendix
\newpage
\section{Pseudocode}
\label{app:pseudocode}
To provide a concise overview of PF-RL, we summarize its training procedure in Algorithm~\ref{algo:pfrl}.
\begin{algorithm}[H]
    \caption{Progress Field Reinforcement Learning (PF-RL)}
    \label{algo:pfrl}
    \begin{flushleft}
        \textbf{Input:} offline dataset $\mathcal D$, VLA policy $\pi_\theta$, projection head $\Phi_\phi$, offline training steps $N_{\mathrm{off}}$, online iterations $N_{\mathrm{on}}$. \\
        \textbf{Initialize:} target projection $\bar{\phi}\leftarrow\phi$; randomly initialize task-level goal prototypes.
    \end{flushleft}
\vspace{-0.15cm}
    \begin{algorithmic}

    \State \textcolor{lightgray}{\# PF-RL Offline Optimization}
    \For{$i=1$ \textbf{to} $N_{\mathrm{off}}$}
        \State Sample $(o_t,\mathbf a_{t:t+H-1}, o_{t^+},o_u,o_g,l)$ from $\mathcal D_\text{offline}$.
        \State Extract VLA representations
        $h \leftarrow \operatorname{sg}[f_{\theta_{\mathrm{VLA}}}(o,l)]$.
        \State Update $\Phi_\phi$ with $\mathcal L_{\mathrm{PF}}$ according to Eq.~\ref{eq:pf_loss}.
        \State Soft-update $\bar{\phi}$ from $\phi$; EMA-update goal prototypes with successful terminal embeddings.
        \State Compute transition-level $A^{\mathrm{prog}}_t$ with the Progress Field according to Eq.~\ref{eq:progress_advantage}.
        \State Update $\pi_\theta$ with the progress-weighted objective in Eq.~\ref{eq:offline_awbc}.
    \EndFor

    \State \textcolor{lightgray}{\# PF-RL Online Fine-Tuning}
    \For{$i=1$ \textbf{to} $N_{\mathrm{on}}$}
        \State Collect online rollout dataset $\mathcal D_\text{online}^{(i)}$ with the current policy $\pi_\theta$.
        \State Sample $(o_t,\mathbf a_{t:t+H-1}, o_{t^+}, o_u,o_g,l)$ from $\mathcal D_\text{online}^{(i)}$.
        \State Extract VLA representations $h \leftarrow \operatorname{sg}[f_{\theta_{\mathrm{VLA}}}(o,l)]$.
        \State Update $\Phi_\phi$ with $\mathcal L_{\mathrm{PF}}$ according to Eq.~\ref{eq:pf_loss}.
        \State Soft-update $\bar{\phi}$ from $\phi$; EMA-update goal prototypes with successful terminal embeddings.
        \State Compute $A^{\mathrm{prog}}_t$ by Eq.~\ref{eq:progress_advantage}, using goal prototypes for failed trajectories.
        \If{\textbf{simulation}}
            \State Combine $A^{\mathrm{GRPO}}$ and $A^{\mathrm{prog}}$ according to Eq.~\ref{eq:sim_advantage}, and update $\pi_\theta$ according to Eq.~\ref{eq:grpo}.
        \Else
            \State Update $\pi_\theta$ with the progress-weighted objective in Eq.~\ref{eq:offline_awbc}.
        \EndIf
    \EndFor

    \State \textbf{return} optimized VLA policy $\pi_\theta$.
    \end{algorithmic}
\end{algorithm}

\section{Experimental Details}
\label{app:exp_details}

We use a shared residual MLP head $\Phi_\phi$ with a hidden dimension of 512 to project final-layer VLA representations of current and goal observations into 32-dimensional embeddings.
Progress Field learning and policy optimization are interleaved throughout offline and online training, updating only the projection head and VLA policy parameters, respectively.
The target head $\Phi_{\bar{\phi}}$ is softly updated after each Progress Field update with a coefficient of $0.005$.
For Progress Field training, $g$ denotes the terminal timestep for both successful and failed trajectories, such that $o_g$ corresponds to the final observation of the trajectory.
The auxiliary observation $o_u$ is sampled from the same trajectory with $t^+<u<g$.
Accordingly, the calibration targets $(g-t)/H$ and $(u-t)/H$ encode observed temporal separations within a trajectory.
For progress-advantage computation, we maintain a task-level goal prototype for each task.
Each prototype is randomly initialized and updated via an exponential moving average of successful terminal embeddings throughout offline and online training.
At each optimization step, a prototype is updated only if the current batch contains successful terminal samples from the corresponding task; otherwise, it remains unchanged.
As successful expert trajectories are available during offline training, the prototypes are progressively anchored to task-completion states before online fine-tuning.
The main LIBERO comparison in Table~\ref{tab:libero_results} and Figure~\ref{fig:learning_curves} is conducted with three independent training seeds, and the full PF-RL results are reused in the ablation studies.
Due to computational constraints, other experiments use a single training seed unless otherwise specified.
We report the checkpoint obtained at the end of each prescribed training stage, without validation-based checkpoint selection.

\subsection{Simulation Settings}
\label{app:sim_details}

For simulation experiments, we instantiate PF-RL with OpenVLA-OFT~\citep{kim2025fine} as the VLA backbone.
For Progress Field learning, we use an expectile coefficient of $\tau=0.7$, a contrastive temperature of $\alpha=0.05$, and an auxiliary subgoal weight of $\lambda_{\mathrm{sub}}=1.0$.
During offline optimization, we use the advantage-weighted objective in Eq.~\ref{eq:offline_awbc} with an advantage temperature of $\beta=0.5$ for 30K steps.
During online reinforcement fine-tuning, we combine the group-relative advantage with the Progress Field advantage using $\lambda_{\mathrm{prog}}=1.0$ and optimize the policy with the clipped GRPO objective.
We perform 100 optimization epochs, with each epoch comprising 64 groups and a group size of 8.
The same rollout data are also used to continuously update the Progress Field head, allowing its value geometry to incorporate states encountered during online exploration.
The Progress Field head and VLA policy are both optimized using AdamW with a micro-batch size of 32.
For LIBERO, we use a learning rate of $2\times10^{-5}$ and an effective batch size of 16,384.
For RoboTwin2.0, we use a learning rate of $2\times10^{-4}$ and an effective batch size of 1,024.
For offline training, we use 30 expert demonstrations per task on LIBERO and 100 per task on RoboTwin2.0, with the latter collected using the official data collection pipeline.
For a fair online comparison, PF-RL and our reproduced RLinf-GRPO baseline are initialized from the same OpenVLA-OFT checkpoint provided by~\citet{li2026simplevla} and trained on the same offline data for 30K steps, using the PF-RL offline objective and standard SFT, respectively.
Online fine-tuning then starts from the resulting policies under the same training budget.
Following~\citet{li2026simplevla}, we evaluate each LIBERO task on 50 held-out test scenarios and each RoboTwin2.0 task on 100 held-out test scenarios.
All simulation experiments are conducted on a cluster of $8$ A800 GPUs.

\subsection{Real-World Settings}
\label{app:real_details}

For real-world experiments, we instantiate PF-RL with $\pi_{0.5}$~\citep{intelligence2025pi_} as the VLA backbone and deploy it on the Agilex Cobot Magic dual-arm platform.
The platform is equipped with two Agilex PIPER robot arms and three RealSense D435 cameras providing one top-down view and two wrist views.
We evaluate PF-RL on four manipulation tasks: \emph{Fold Clothes}, \emph{Stack all Plates}, \emph{Put all Blocks into Bucket}, and \emph{Fill Vase and Arrange Flowers}.
We treat \emph{Fold Clothes} as a single-task setting, while the latter three tasks are jointly trained under a multi-task setting.
For offline training, we collect 200 expert demonstrations for \emph{Fold Clothes} and 100 demonstrations per task for the multi-task setting.
We first perform offline PF-RL using the advantage-weighted objective in Eq.~\ref{eq:offline_awbc} with an advantage temperature of $\beta=1.5$, optimizing the policy for 20K steps with a batch size of 64.
Compared with simulation, we use a larger $\beta$ to reduce aggressive advantage reweighting and encourage more conservative policy updates in the more complex real-world setting.
Starting from the resulting offline policy, we further perform real-world online PF-RL.
To improve the safety and efficiency of real-world exploration, we employ human-in-the-loop intervention~\citep{evorl2026} during online interaction.
A human operator monitors policy execution and temporarily takes control when the robot enters an undesirable state, fails to make progress, or requires correction for recovery.
Control is returned to the policy once the robot reaches a recoverable state.
The intervention segments are retained as part of the collected trajectories and used for subsequent policy and Progress Field updates.
Interventions are triggered as needed during policy execution.
We conduct two RL iterations and collect 20 rollouts per task in each iteration, following the setting of $\pi^*_{0.6}$~\citep{intelligence2025pi06vlalearnsexperience}.
The newly collected trajectories are then evaluated using the Progress Field advantage and used to update the policy and the Progress Field head.
All other Progress Field architecture and training hyperparameters follow the simulation setting.
All real-world training is conducted on a cluster of $8$ A800 GPUs, while policy inference is performed on a single A800 GPU.

\begin{figure*}[t]
    \centering
    \includegraphics[width=\linewidth]{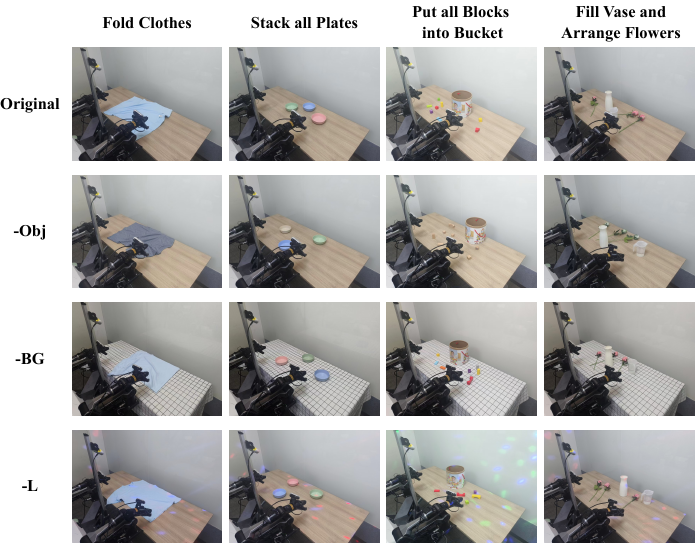}
    \caption{
    Overview of real-world tasks and visual variations.
    Columns show the four manipulation tasks: \emph{Fold Clothes}, 
    \emph{Stack all Plates}, \emph{Put all Blocks into Bucket}, and 
    \emph{Fill Vase and Arrange Flowers}.
    Rows correspond to the original environment settings and three visual variations involving object~(-Obj), background~(-BG), and lighting~(-L) changes.
    }
    \label{fig:real_world_task_overview}
    \vspace{-15pt}
\end{figure*}

\section{Real-World Tasks and Visual Variations}
\label{app:real_world_tasks}

Figure~\ref{fig:real_world_task_overview} provides an overview of the four real-world manipulation tasks and the visual variations used for evaluation.
The tasks cover diverse long-horizon manipulation behaviors, including cloth folding, plate stacking, multi-block collection, and vase arrangement.
For each task, we evaluate the policy in the Original setting and under three visual variations while keeping the task objective unchanged.
Specifically, (-Obj) replaces the task-relevant objects with alternative instances, (-BG) changes the visual background of the workspace, and (-L) introduces substantial lighting variations.
Since these visual variations are not observed during training, they provide a direct evaluation of the robustness of the VLA policy to out-of-distribution visual changes.

\section{Additional Analysis of Progress Modeling}
\label{app:progress_modeling}

\subsection{Experimental Setup for Progress Modeling Comparison}
\label{app:progress_modeling_setup}

For the direct scalar prediction baseline in Section~\ref{sec:exp_ablation}, we replace the Progress Field with a lightweight prediction head that directly regresses a goal-conditioned scalar value from the current and goal VLA representations.
The scalar baseline uses the Bellman-consistency and step-to-goal calibration objectives without goal contrastive learning, while the VLA backbone, policy optimization procedure, and training budget remain unchanged.
Thus, the primary difference lies in whether progress is modeled through direct scalar prediction or goal-conditioned value geometry.
For the held-out-task experiments in Figure~\ref{fig:progress_modeling_analysis}, both methods start from the same pretrained SFT VLA policy from~\citet{li2026simplevla}.
We train the corresponding progress head on nine tasks from each LIBERO suite, while excluding the evaluated task from progress-model training.
Since the held-out task is excluded from prior progress-model training, no task-specific goal reference is available beforehand.
We therefore initialize it from successful rollouts collected by the policy at the start of online fine-tuning.
For direct scalar prediction, we average the terminal VLA representations to obtain the goal input, whereas for Progress Field, we average the corresponding projected embeddings to construct the goal prototype.
During online adaptation, the corresponding progress head continues to be updated following the standard PF-RL training procedure.

\subsection{Additional Comparison with Scalar Prediction Variants}
\label{app:scalar_prediction_variants}
To further evaluate our geometric value parameterization, we introduce an MLP readout variant that replaces the distance-based value computation with a scalar prediction head $\Phi_{\mathrm{scalar}}(z_t,z_g)$ feeding on the projected embeddings \(z_t\) and \(z_g\), while retaining the projection architecture and training objectives of Progress Field.
The projection head and scalar readout are jointly trained, with the same goal contrastive loss applied to the goal Progress Field embeddings.
Table~\ref{tab:scalar_prediction_variants} compares this variant with direct scalar prediction from VLA latent representation, PF-RL without $\mathcal{L}_{\mathrm{con}}$, and full PF-RL.
All of the variants share the same reinforcement fine-tuning pipeline, with differences limited to the value parameterization and the inclusion of goal contrastive learning.
PF-RL outperforms both scalar prediction variants across all four suites. 
Even without $\mathcal{L}_{\mathrm{con}}$, PF-RL outperforms direct scalar prediction. These results support the effectiveness of geometric value parameterization, with further gains from goal contrastive learning.
\begin{table}[t]
    \centering
    \caption{
        Additional comparison with scalar prediction variants on LIBERO.
    }
    \label{tab:scalar_prediction_variants}
    \small
    \setlength{\tabcolsep}{4pt}
    \renewcommand{\arraystretch}{1.1}
    \begin{tabular}{@{}lccccccc@{}}
        \toprule
        \textbf{Variant}
        & \textbf{Value Estimation}
        & $\mathcal{L}_{\mathrm{con}}$
        & \textbf{Spatial}
        & \textbf{Object}
        & \textbf{Goal}
        & \textbf{Long}
        & \textbf{Avg.} \\
        \midrule
        Direct scalar prediction
        & $\Phi_{\mathrm{scalar}}(h_t,h_g)$
        & $\times$
        & 97.0 & 97.4 & 95.6 & 95.8 & 96.5 \\
        PF-RL w/o $\mathcal{L}_{\mathrm{con}}$
        & $-\lVert z_t-z_g\rVert_2$
        & $\times$        
        & 98.4 & 98.4 & 98.0 & 97.6  & 98.1 \\
        \midrule
        MLP readout variant
        & $\Phi_{\mathrm{scalar}}(z_t,z_g)$
        & $\checkmark$
        & 98.8 & 98.2 & 94.4 & 95.2 & 96.7 \\
        PF-RL
        & $-\lVert z_t-z_g\rVert_2$
        & $\checkmark$
        & \textbf{99.2} & \textbf{99.5} & \textbf{98.8} & \textbf{99.0} & \textbf{99.1} \\
        \bottomrule
    \end{tabular}
\end{table}

\begin{figure*}[t]
    \centering

    \begin{tabular}{@{}cccc@{}}

        \includegraphics[width=0.24\textwidth]
        {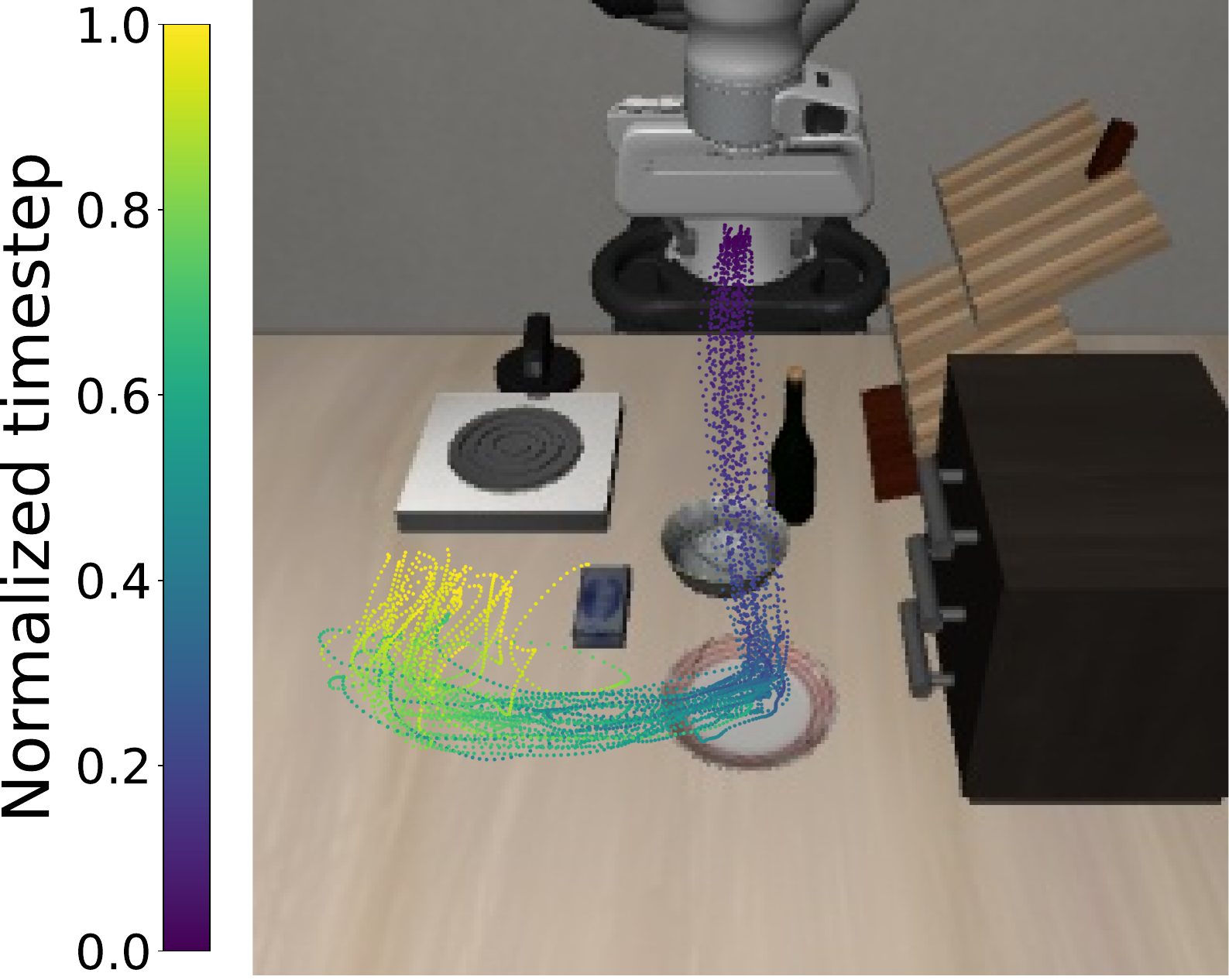}
        &
        \includegraphics[width=0.21\textwidth]
        {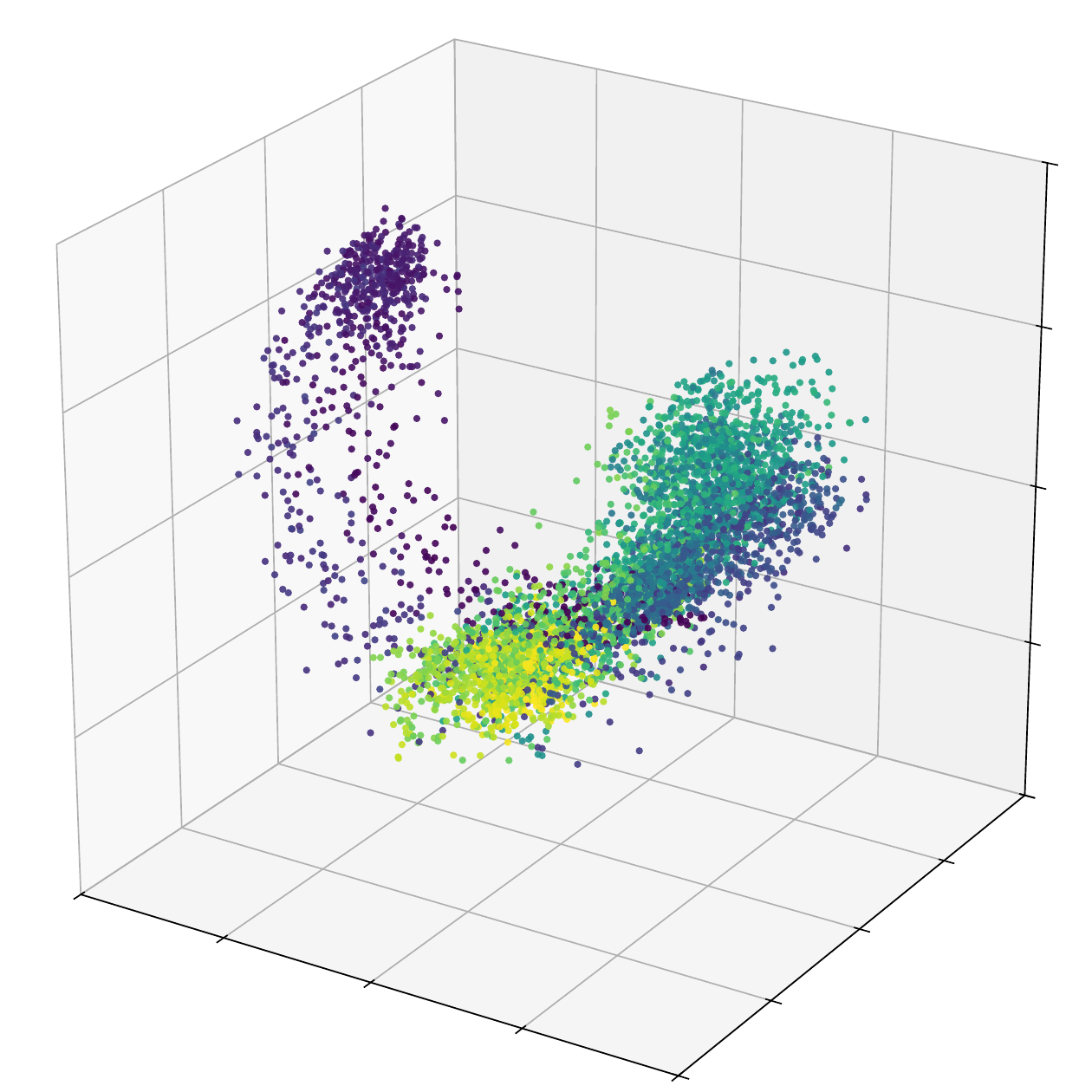}
        &
        \includegraphics[width=0.21\textwidth]
        {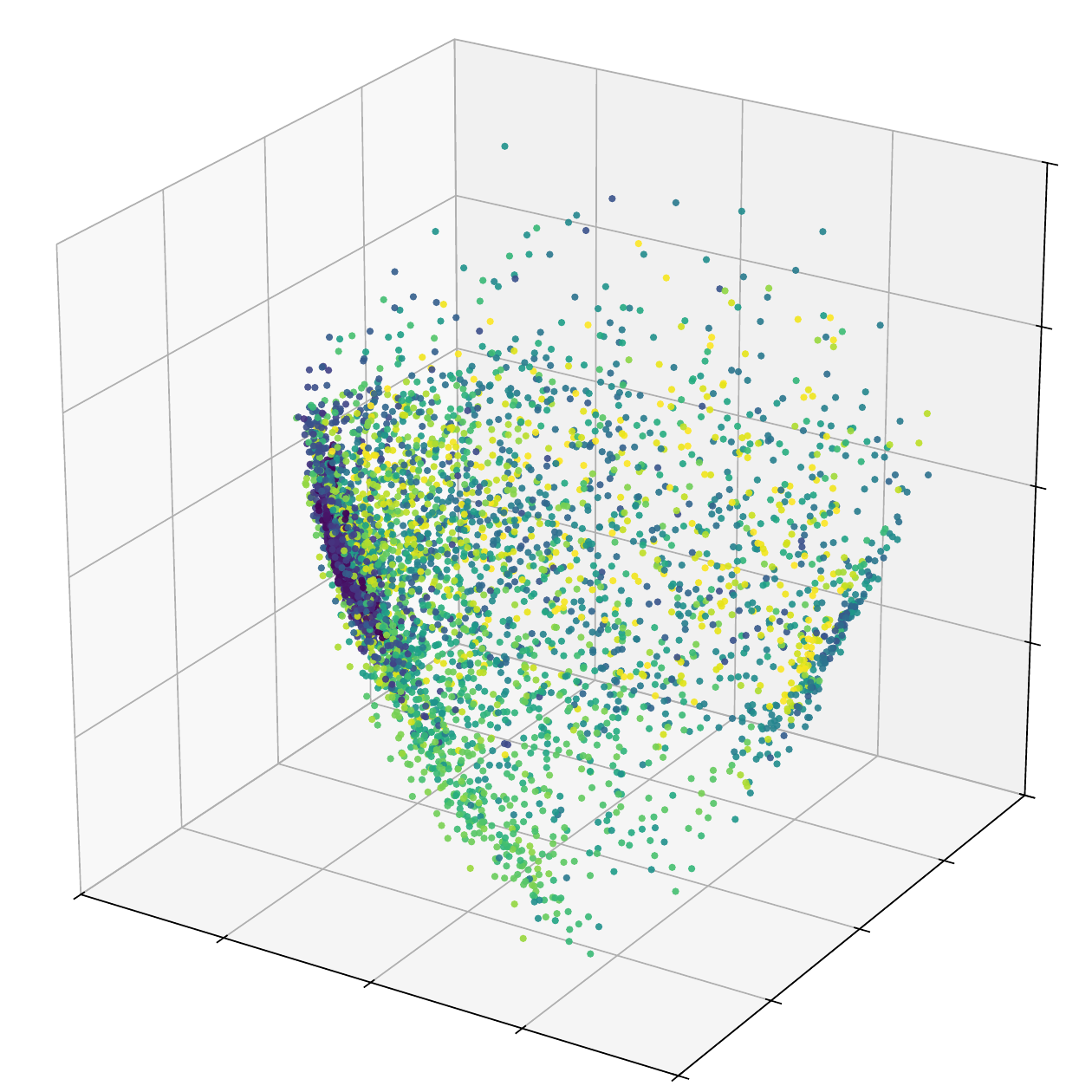}
        &
        \includegraphics[width=0.21\textwidth]
        {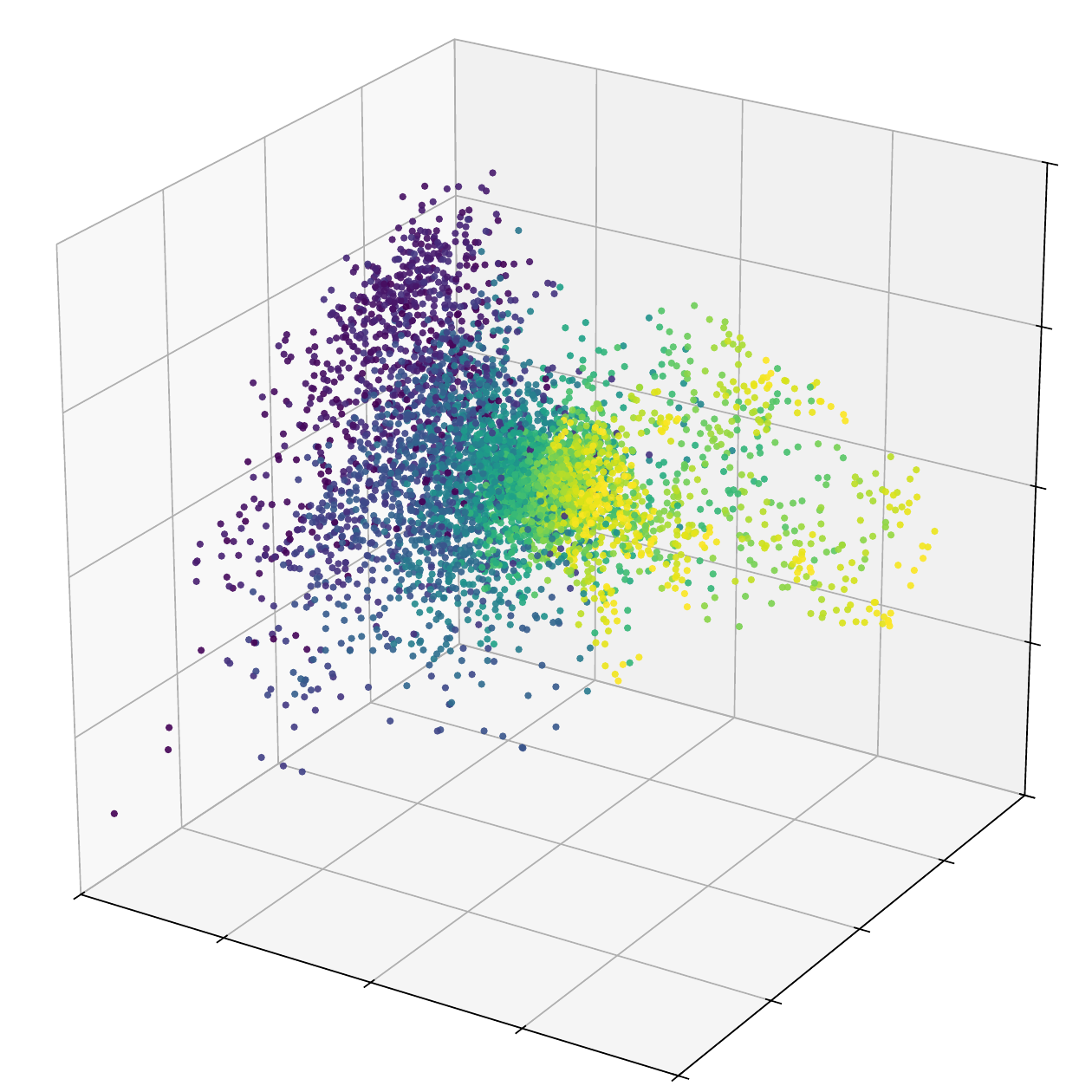}
        \\[3pt]

        \includegraphics[width=0.24\textwidth]
        {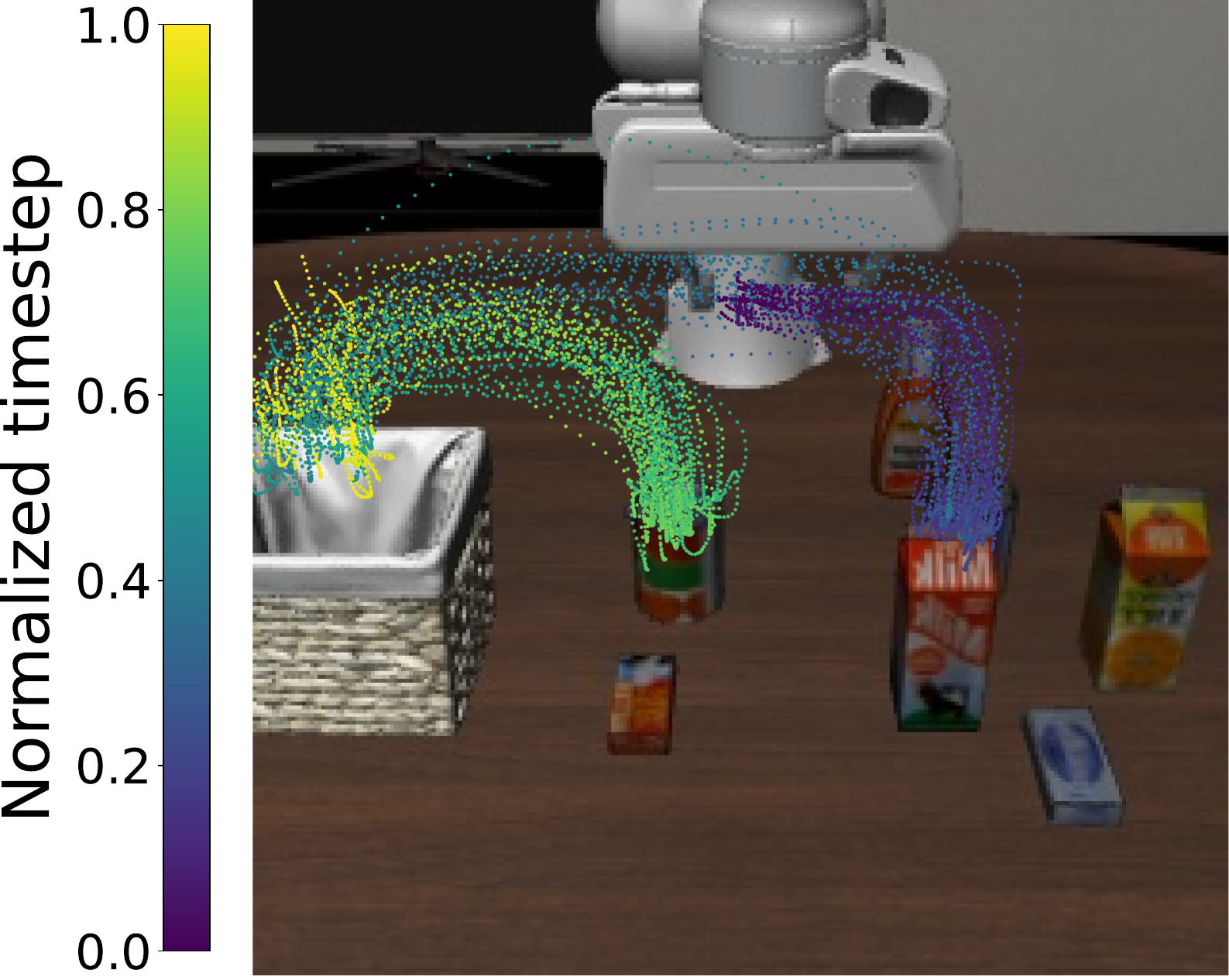}
        &
        \includegraphics[width=0.21\textwidth]
        {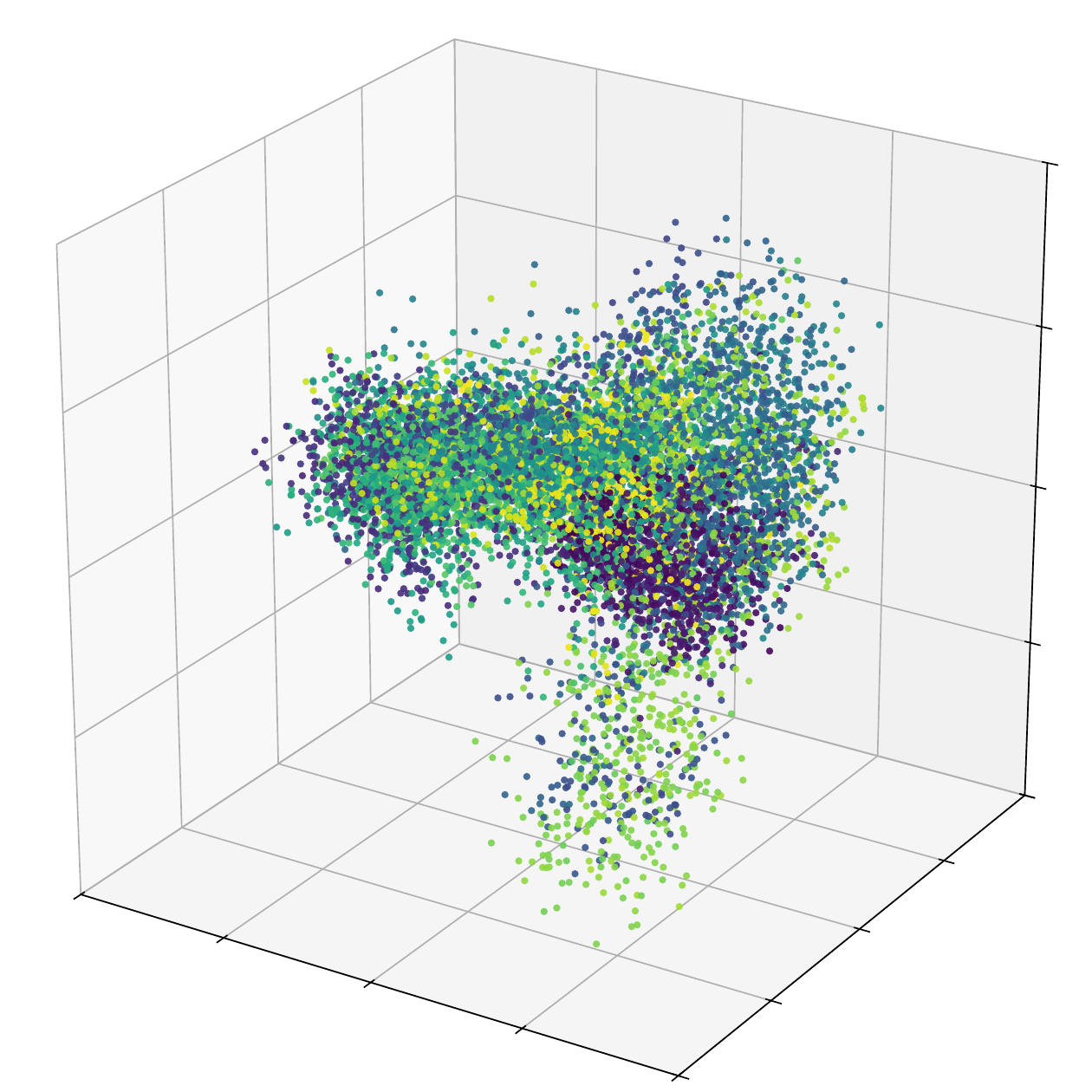}
        &
        \includegraphics[width=0.21\textwidth]
        {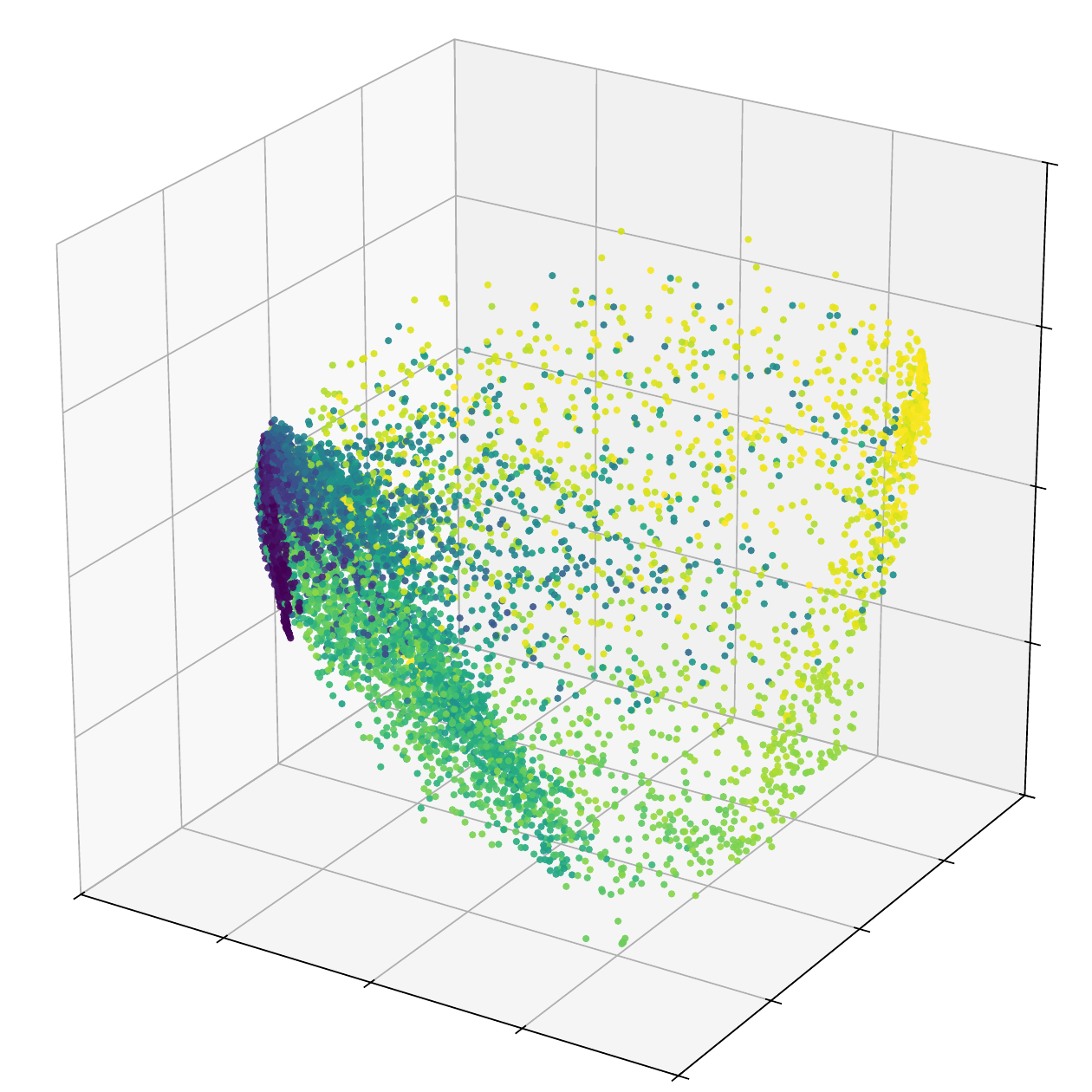}
        &
        \includegraphics[width=0.21\textwidth]
        {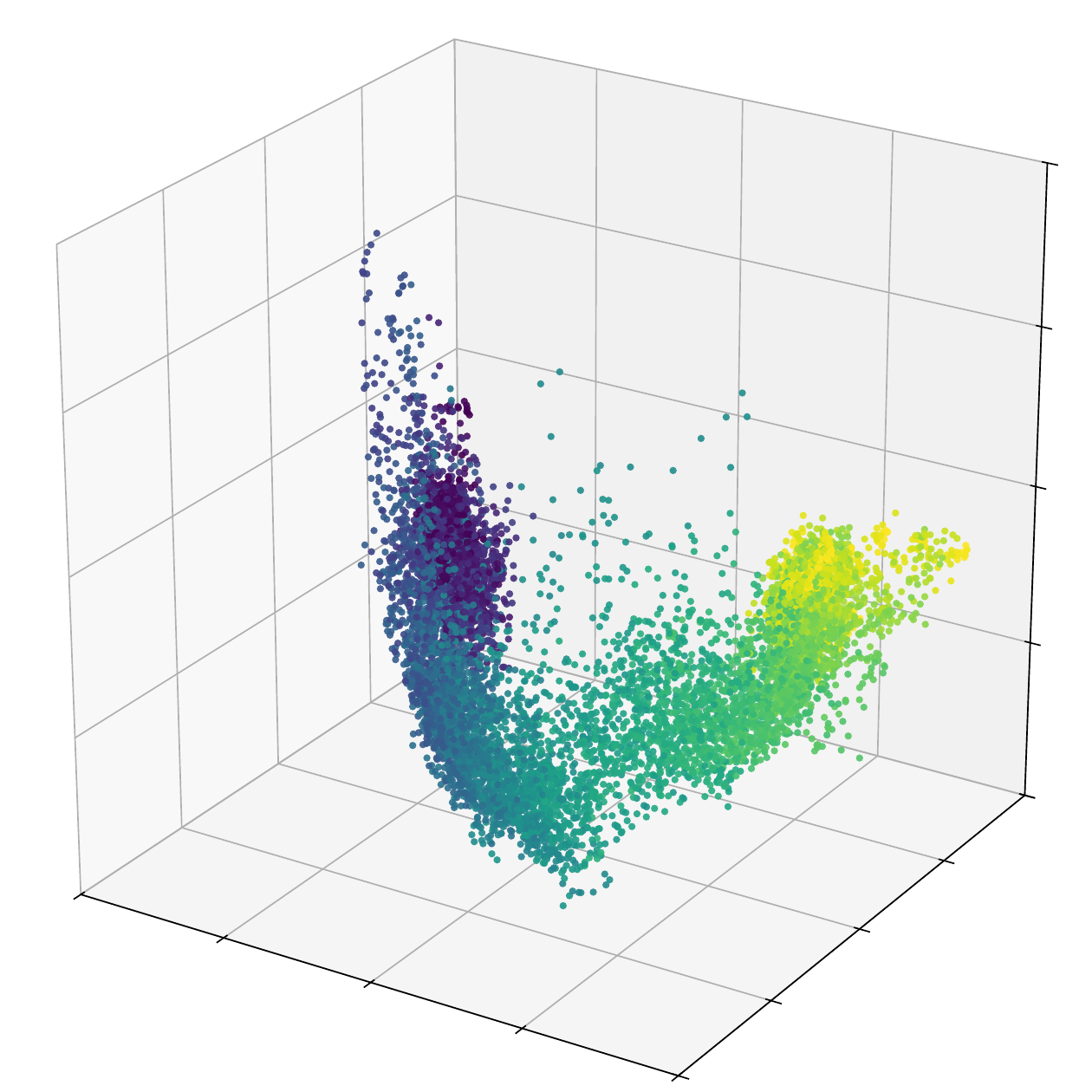}
        \\[6pt]

        (a) Task progression
        &
        (b) Raw VLA space
        &
        (c) Scalar prediction
        &
        (d) Progress Field

    \end{tabular}

        \caption{
        Representation visualization on held-out tasks.
        The top row shows LIBERO-Goal Task 5,
        \textit{push the plate to the front of the stove},
        while the bottom row shows LIBERO-Long Task 0,
        \textit{put both the alphabet soup and the tomato sauce in the basket}.
        (a)~Expert end-effector trajectories,
        (b)~Raw VLA representations,
        (c)~Intermediate representations of the direct scalar prediction head,
        and (d)~Projected embeddings learned by Progress Field.
    }
    \label{fig:heldout_representation_vis}
\end{figure*}
\subsection{Representation Visualization on Held-out Tasks}
\label{app:heldout_representation}

To further evaluate the learned progress representations, Figure~\ref{fig:heldout_representation_vis} visualizes representations along expert trajectories from the same held-out tasks, using the progress models before online adaptation.
Although direct scalar prediction captures some progress-related structure, its representations remain relatively dispersed with substantial overlap across trajectory stages.
In contrast, the Progress Field exhibits a clearer and more coherent progression from early to late stages, despite these tasks being excluded from progress-model training.
This representation-level evidence further suggests that the learned goal-conditioned geometry captures progress structure that transfers more effectively across related tasks.

\section{Additional Ablation Results}
\label{app:additional_ablation}

\subsection{Auxiliary Subgoal Training}
\label{app:additional_ablation_subgoal}

\begin{wrapfigure}{r}{0.44\linewidth}
    \centering
    \vspace{-10pt}
    \includegraphics[width=\linewidth]
    {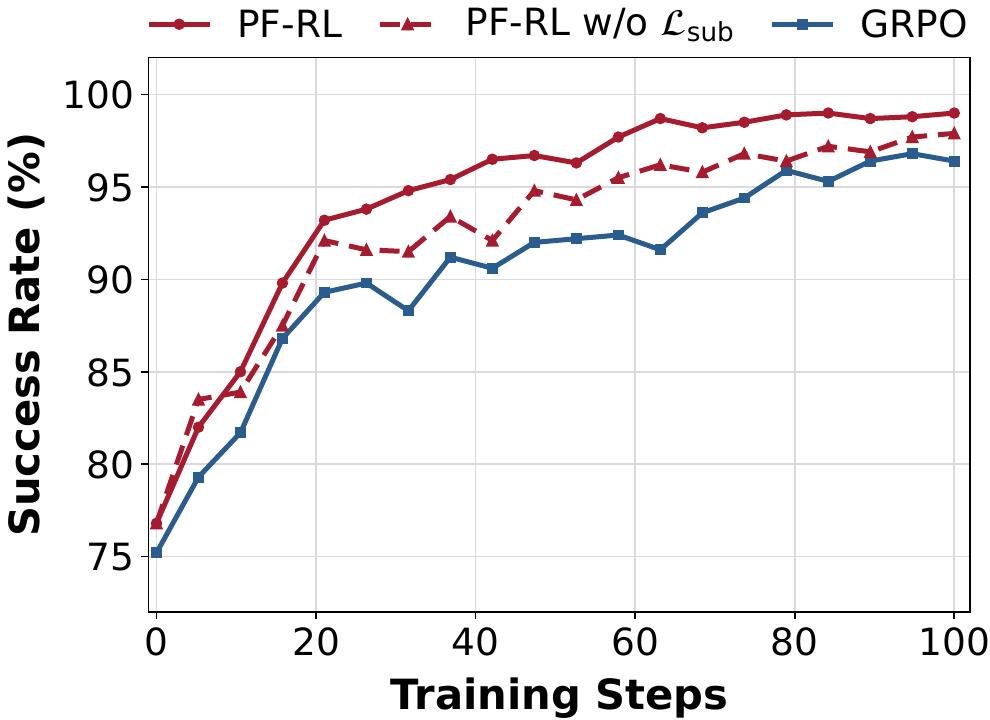}
    \vspace{-15pt}
    \caption{Auxiliary subgoal ablation.}
    \label{fig:subgoal_ablation}
    \vspace{-12pt} 
\end{wrapfigure}
\paragraph{Effect of Auxiliary Subgoal Training.}
Figure~\ref{fig:subgoal_ablation} compares PF-RL with vanilla GRPO and a variant trained without the auxiliary subgoal objective $\mathcal L_{\mathrm{sub}}$ on LIBERO-Long.
The results show that PF-RL outperforms both vanilla GRPO and the terminal-goal-only variant during training on LIBERO-Long. 
Even without $\mathcal{L}_{\mathrm{sub}}$, PF-RL still improves over GRPO, demonstrating that goal-conditioned progress guidance alone provides an effective learning signal for long-horizon VLA policy optimization.
However, this variant underperforms the full PF-RL method and exhibits more pronounced fluctuations throughout training.
In contrast, introducing intermediate observations as auxiliary subgoals more fully exploits the intermediate structure of long-horizon rollout trajectories, thereby accelerating convergence and yielding a higher success rate under the same training budget. 
These results suggest that auxiliary subgoal training better exploits intermediate structure in long-horizon rollout trajectories, enabling further policy improvement beyond terminal-goal supervision alone.

\subsection{Progress Field Training Objectives.}
\label{app:additional_ablation_objective}
Table~\ref{tab:pf_objective_ablation} summarizes the Progress Field objective ablations corresponding to Figure~\ref{fig:tpae_loss_ablation}.
The full objective performs best across all four LIBERO suites, with removing $\mathcal{L}_{\mathrm{td}}$ causing the largest degradation on LIBERO-Long.
The TD-only metric variant adopts HILP-style value learning~\citep{park2024foundation} within the same PF-RL pipeline, retaining both terminal-goal and auxiliary-subgoal TD losses.
Its lower performance, particularly on LIBERO-Goal and LIBERO-Long, supports the additional benefit of step-to-goal calibration and goal contrastive learning beyond metric TD learning alone.
These results support combining Bellman consistency, step-to-goal calibration, and goal contrastive learning for effective progress-aware policy optimization.

\begin{table}[t]
    \centering
    \caption{Results of the Progress Field training objective ablation on LIBERO.}
    \label{tab:pf_objective_ablation}
    \small
    \begin{tabular}{lccccc}
        \toprule
        \textbf{Method}
        & \textbf{Spatial}
        & \textbf{Object}
        & \textbf{Goal}
        & \textbf{Long}
        & \textbf{Avg.} \\
        \midrule

        PF-RL w/o $\mathcal{L}_{\mathrm{td}}$
        & 97.8
        & 98.8
        & 96.6
        & 94.4
        & 96.9 \\

        PF-RL w/o $\mathcal{L}_{\mathrm{step}}$
        & 97.4
        & 98.4
        & 96.0
        & 95.2
        & 96.8 \\

        PF-RL w/o $\mathcal{L}_{\mathrm{con}}$
        
        & 98.4
        & 98.4
        & 98.0
        & 97.6
        & 98.1 \\
        PF-RL w/o \(\mathcal L_{\mathrm{step}}\) and \(\mathcal L_{\mathrm{con}}\)~(TD-only)
        & 97.6
        & 99.0
        & 95.4
        & 95.6
        & 96.9
        \\\midrule
                PF-RL
        & \textbf{99.2}
        & \textbf{99.5}
        & \textbf{98.8}
        & \textbf{99.0}
        & \textbf{99.1} \\

        \bottomrule
    \end{tabular}
\end{table}

\subsection{Analysis of Progress-Aware Advantage}

We further analyze the Progress Field advantage in online policy optimization.
Recall that PF-RL combines the trajectory-level group-relative advantage with the transition-level progress advantage as
$A^{\mathrm{sim}}_{i,t} = A^{\mathrm{GRPO}}_i + \lambda_{\mathrm{prog}} A^{\mathrm{prog}}_{i,t}$.
We evaluate PF-RL across a range of $\lambda_{\mathrm{prog}}$ values on all four LIBERO suites.
As shown in Table~\ref{tab:prog_adv_ablation}, PF-RL consistently outperforms the original GRPO baseline across a broad range of $\lambda_{\mathrm{prog}}$, validating the benefit of incorporating progress-aware credit into online policy optimization.
Notably, $\lambda_{\mathrm{prog}}=1.0$ yields the strongest average performance across the four suites, suggesting that appropriately combining trajectory-level group-relative advantages with transition-level progress advantages benefits online policy optimization.
We further remove $A^{\mathrm{GRPO}}$ and optimize VLA using only $A^{\mathrm{prog}}$.
Using $A^{\mathrm{prog}}$ alone remains competitive, while adding $A^{\mathrm{GRPO}}$ further improves performance through complementary trajectory-level feedback.

\begin{table}[t]
    \centering
    \caption{Analysis of the progress advantage on LIBERO.}
    \label{tab:prog_adv_ablation}
    \small
    \begin{tabular}{lccccc}
        \toprule
        \textbf{Setting}
        & \textbf{Spatial}
        & \textbf{Object}
        & \textbf{Goal}
        & \textbf{Long}
        & \textbf{Avg.} \\
        \midrule
        GRPO ($\lambda_{\mathrm{prog}}=0$)
        & 97.6 & 98.7 & 95.2 & 96.8 & 97.1 \\
        PF-RL ($\lambda_{\mathrm{prog}}=0.1$)
        & 98.8 & 98.8 & 98.0 & 97.2 & 98.2 \\
        PF-RL ($\lambda_{\mathrm{prog}}=0.5$)
        & 98.6 & \textbf{99.8}  & 97.6 & 98.2 & 98.6 \\
        \textbf{PF-RL ($\lambda_{\mathrm{prog}}=1.0$)}
        & \textbf{99.2} & \textbf{99.5} & \textbf{98.8} & 99.0 & \textbf{99.1} \\
        PF-RL ($\lambda_{\mathrm{prog}}=2.0$)
        & 99.0 & 99.4 & 98.0 & \textbf{99.2} & 98.9 \\
        PF-RL ($\lambda_{\mathrm{prog}}=5.0$)
        & 98.8 & 99.2 & 96.2 & 97.2 & 97.9 \\
        PF-RL ($A^{\mathrm{prog}}$ only)
        & 97.8 & 99.2 & 97.4 & 96.4 & 97.7 \\
        \bottomrule
    \end{tabular}
\end{table}

\begin{figure}[t]
    \centering

    \begin{subfigure}[t]{0.42\linewidth}
        \centering
        \includegraphics[width=\linewidth]{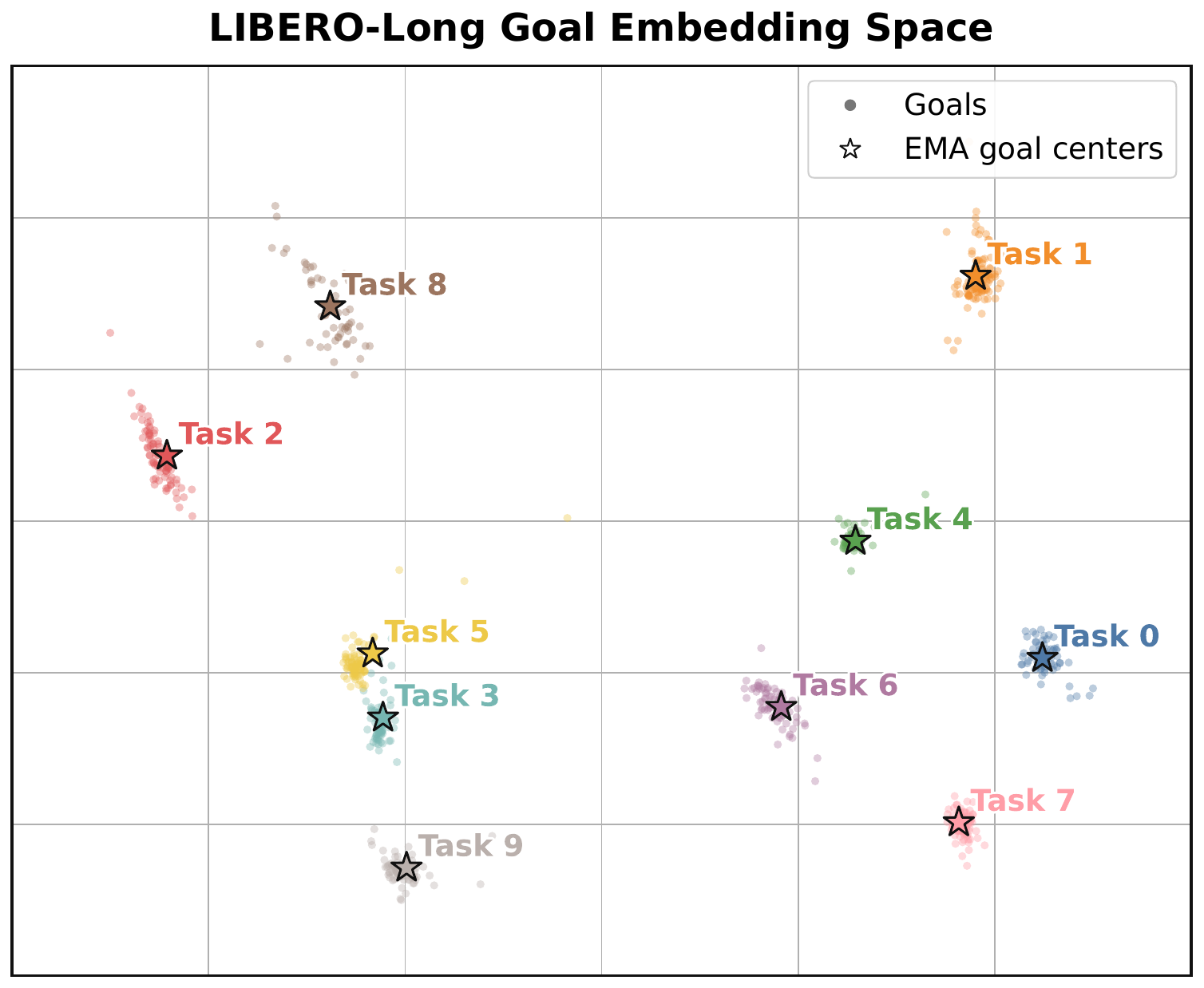}
        \caption{LIBERO-Long}
        \label{fig:goal_embedding_libero}
    \end{subfigure}
    \hfill
    \begin{subfigure}[t]{0.42\linewidth}
        \centering
        \includegraphics[width=\linewidth]{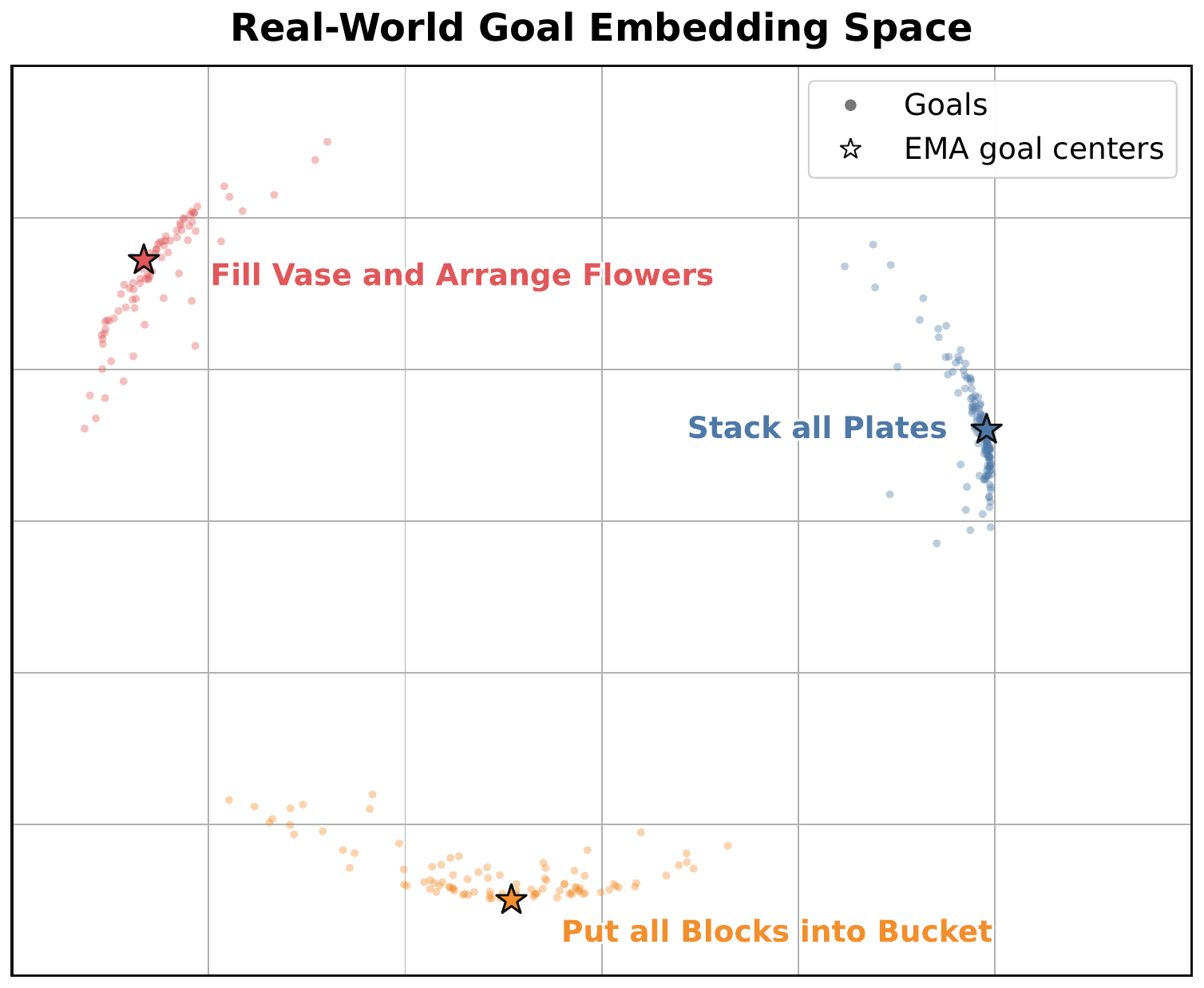}
        \caption{Real-world tasks}
        \label{fig:goal_embedding_real}
    \end{subfigure}

    \caption{PCA visualization of goal embeddings from successful trajectories, colored by task.}
    \vspace{-2pt}
    \label{fig:goal_embedding_vis}
\end{figure}

\section{Visualization of Goal Embedding Space}
\label{app:goal_embedding_vis}
To further examine the structure learned in the goal embedding space, we visualize goal embeddings from all tasks within the LIBERO-Long suite and the real-world multi-task settings.
Specifically, we collect the goal embeddings of successful trajectories for each task and project them into two dimensions using PCA, with different tasks shown in different colors.
As shown in Figure~\ref{fig:goal_embedding_vis}, goal embeddings from the same task form compact clusters, while those from different tasks are clearly separated, indicating that the learned Progress Field preserves task-specific goal semantics.
We further mark the corresponding per-task goal prototypes maintained in the goal embedding table with stars.
These prototypes lie close to their respective task clusters, suggesting that the prototype provides a representative reference for evaluating task progress, particularly for failed trajectories where the terminal observation does not correspond to task completion.

\section{Analysis of Progress Advantages}
\begin{wrapfigure}{r}{0.38\linewidth}
\vspace{-15pt}
    \centering
    \includegraphics[width=\linewidth]
    {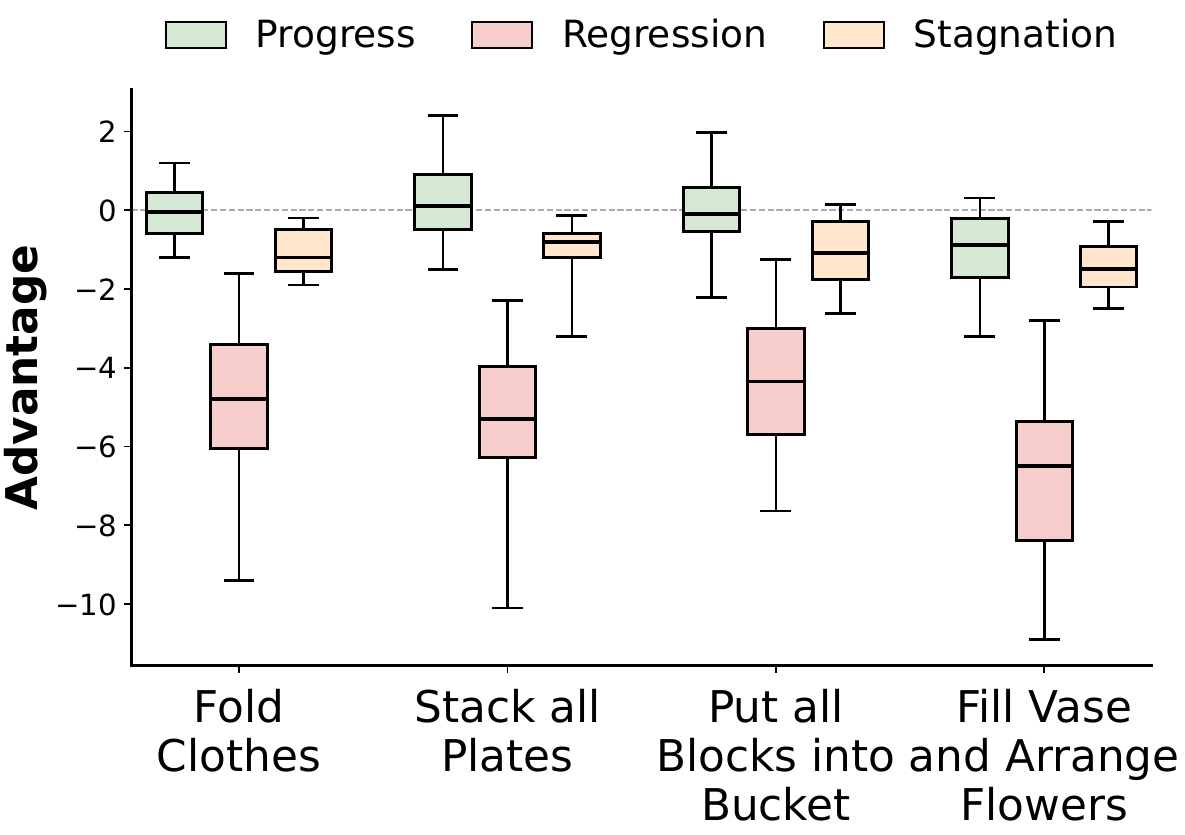}
    \vspace{-18pt}
    \caption{Progress advantage distributions across execution behaviors.}
    \label{fig:adv_box}
    \vspace{-13pt}
\end{wrapfigure}
We further examine the progress advantages induced by the learned Progress Field under different execution behaviors.
Specifically, we manually annotate real-world rollout segments as \emph{progress}, \emph{regression}, or \emph{stagnation}, according to whether the robot advances toward the goal, loses progress, or struggles to advance.
For each of the four tasks, we collect 10 segments per category and average $A_t^{\mathrm{prog}}$ over action chunks within each segment.
Figure~\ref{fig:adv_box} shows the highest median advantages for progress, followed by stagnation and regression, across all tasks.
Progress segments may receive near-zero or negative advantages because the signal evaluates progress relative to the learned temporal scale.
These results indicate that progress advantages faithfully reflect local task progress, supporting their use as reliable feedback for policy optimization.

\section{Progress Field Visualizations}
\label{app:pf_visualizations}

\begin{figure*}[t]
    \centering
    \includegraphics[width=\linewidth]
    {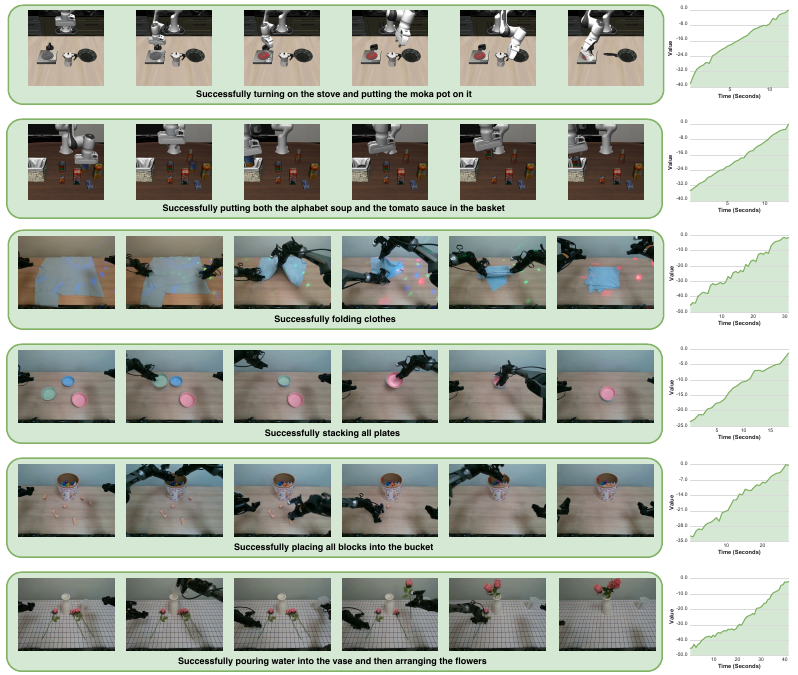}
    \caption{Successful rollout trajectories and corresponding goal-conditioned values.}
    \label{fig:pf_success_vis}
\end{figure*}

\begin{figure*}[t]
    \centering
    \includegraphics[width=\linewidth]
    {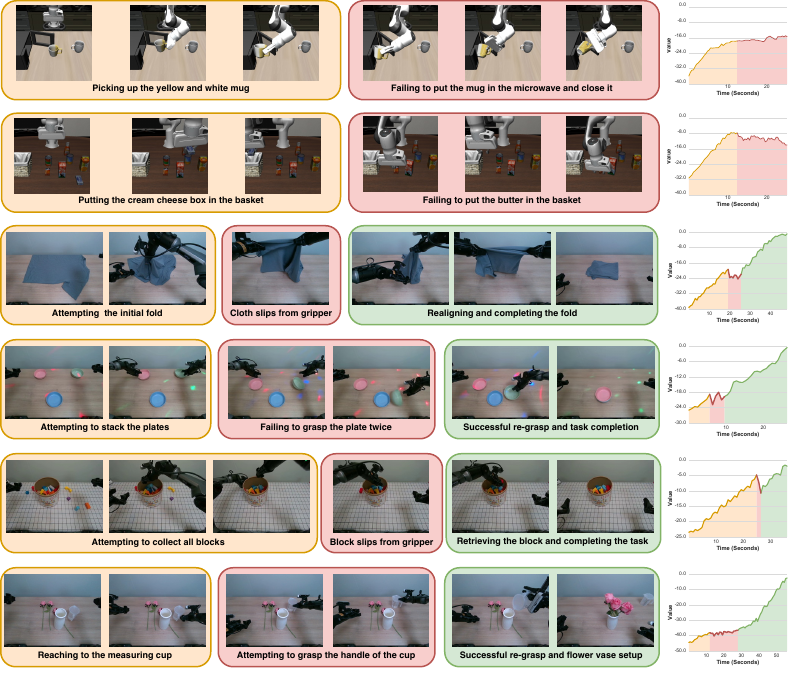}
    \vspace{-5pt}
    \caption{Rollout trajectories with intermediate failures and corresponding values.}
    \label{fig:pf_recovery_vis}
\end{figure*}

Figure~\ref{fig:pf_success_vis} and Figure~\ref{fig:pf_recovery_vis} provide visualizations of the inferred trajectories as well as the corresponding goal-conditioned values induced by the Progress Field.
Figure~\ref{fig:pf_success_vis} shows successful executions across simulation tasks and four real-world tasks with different variations.
Despite their different manipulation behaviors and execution horizons, the estimated values exhibit a consistent overall increase as the robot progresses toward task completion, approaching higher values as the goal is reached.
Figure~\ref{fig:pf_recovery_vis} further examines trajectories that encounter intermediate execution failures.
In simulation, when the robot struggles to make progress toward the final goal, the estimated progress value plateaus or even declines over subsequent timesteps.
Across different failure modes, including unsuccessful grasps and object slippage, the estimated value first decreases or plateaus, and increases again once the robot resumes progress toward the goal.
This indicates that the learned value tracks task progress rather than merely elapsed execution time.
These visualizations show that the Progress Field captures both successful task advancement and temporary execution degradation, providing meaningful transition-level signals for progress-aware policy optimization.


\newpage

\section{Computational Overhead}
\label{app:computational_overhead}
\begin{wrapfigure}{r}{0.35\linewidth}
\vspace{-15pt}
    \centering
    \includegraphics[width=\linewidth]
    {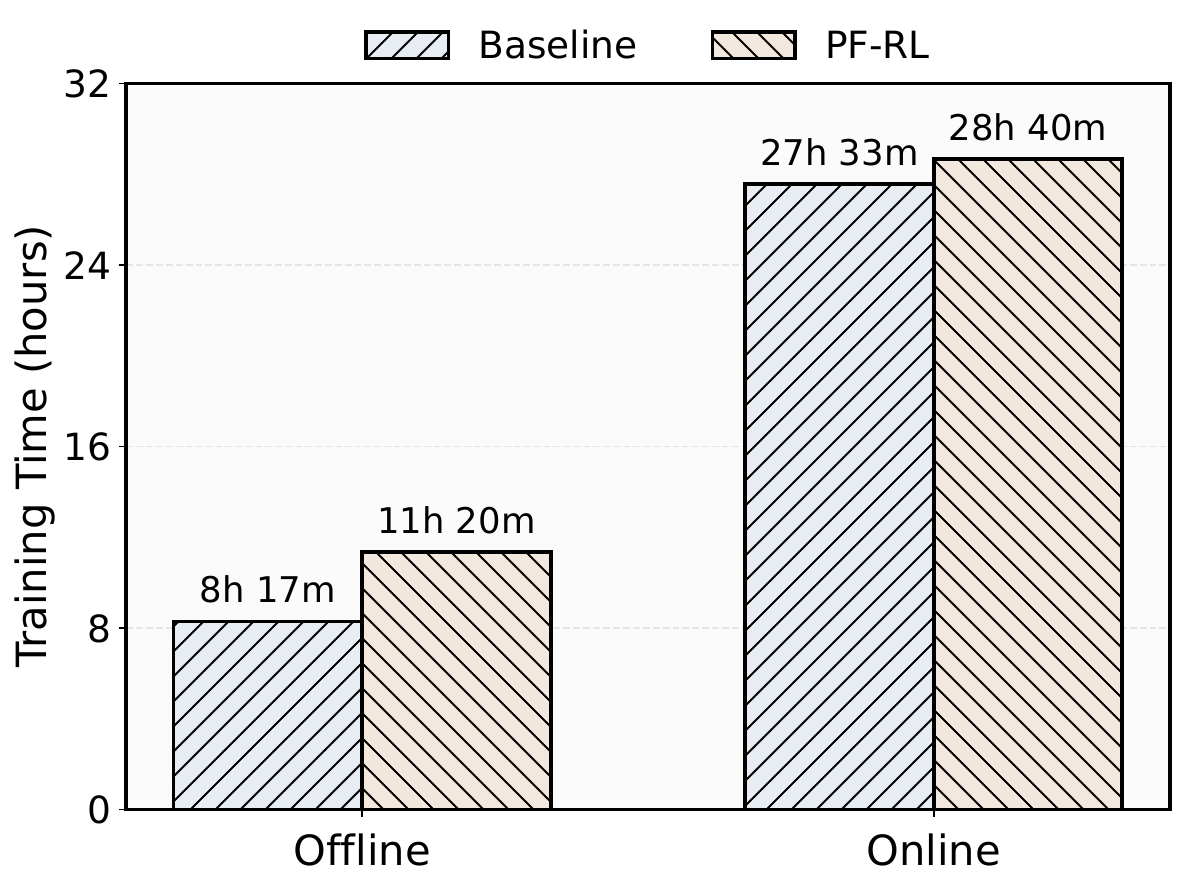}
    \vspace{-20pt}
    \caption{Training time of baselines and PF-RL.}
    \label{fig:training_time_comparison}
    \vspace{-13pt}
\end{wrapfigure}
To ensure a fair comparison, we measure the overhead of PF-RL against the offline SFT and online GRPO baselines on a cluster of 8 A800 GPUs, 30K offline optimization steps following \citet{li2026simplevla}, and 100 online optimization epochs.
As shown in Figure~\ref{fig:training_time_comparison}, offline training time increases from 8.3 to 11.3 hours, mainly due to additional encoding of goal observations, despite stopping Progress Field gradients at the VLA representations.
In contrast, online fine-tuning increases only from 27.6 to 28.7 hours.
A simple caching strategy largely enables this low overhead: we store the VLA latent representations during online rollout collection and reuse them for subsequent progress advantage computation, avoiding additional forward passes through the VLA backbone.
The remaining online overhead mainly comes from updating the
lightweight Progress Field head.

\section{Limitations}
\label{app:limitations}
PF-RL relies on successful trajectories in the offline dataset to construct initial task-level goal references.
Although failed trajectories also contribute to Progress Field learning, an offline dataset containing only failed trajectories would not provide the successful terminal embeddings needed to establish this table.
Extending PF-RL to such failure-only offline settings remains future work.

\end{document}